%% file: top.tex
\documentclass[10pt,twocolumn,letterpaper]{article}

\usepackage{cvpr}
\usepackage{times}
\usepackage{epsfig}
\usepackage{graphicx}
\usepackage{amsmath}
\usepackage{amssymb}
\usepackage{amsfonts}
\usepackage{subcaption}
\usepackage{tabularx}
\usepackage{bbm}
\usepackage[affil-it]{authblk}
\usepackage{array}
\usepackage{url}

\usepackage{mathtools}
\usepackage{multirow}

\input{defs}

\usepackage{booktabs}
\usepackage[pagebackref=true,breaklinks=true,letterpaper=true,colorlinks,bookmarks=false]{hyperref}

 \cvprfinalcopy 
\def\cvprPaperID{***} 

\ifcvprfinal\fi

\begin{document}

\sloppy

\title{Real-Time Camera Pose Estimation for Sports Fields}


\newcommand\CoAuthorMark{\footnotemark[\arabic{footnote}]}

\author[1]{Leonardo Citraro\thanks{Contributed equally.}}
\author[2]{Pablo M\'arquez-Neila\protect\CoAuthorMark{}}
\author[1]{Stefano Savar\`e}
\author[3]{Vivek Jayaram}
\author[3]{Charles Dubout}
\author[3]{F\'elix Renaut}
\author[3]{Andr\'es Hasfura}
\author[3]{Horesh Ben Shitrit}
\author[1]{Pascal Fua}
\affil[1]{Computer Vision Laboratory, \'{E}cole Polytechnique F\'{e}d\'{e}rale de Lausanne}
\affil[2]{ARTORG Center for Computer Aided Surgery, University of Bern  }
\affil[3]{Second Spectrum Inc.}

\maketitle

\input{0_abstract.tex}
\input{1_introduction.tex}
\input{2_related_work.tex}
\input{3_method.tex}
\input{4_experiments.tex}

\input{5_conclusion.tex}
\input{7_acknowledgments.tex}

\appendix
\input{6_appendix.tex}


\end{document}

%% file: defs.tex
\usepackage{amsmath}
\usepackage{amssymb}
\usepackage[utf8]{inputenc}
\usepackage[T1]{fontenc}
\usepackage{algorithm}
\usepackage{adjustbox}
\usepackage{multirow}
\usepackage[noend]{algpseudocode}
\usepackage{graphicx}
\usepackage{color}
\usepackage{booktabs, multirow}
\usepackage{wrapfig}
\usepackage{cite}



\graphicspath{{images/}}

\def\BState{\State\hskip-\ALG@thistlm}

\newcommand{\comment}[1]{}
\newcommand{\parag}[1]{\vspace{-1mm}\paragraph{#1}}

\newcommand\blfootnote[1]{%
  \begingroup
  \renewcommand\thefootnote{}\footnote{#1}%
  \addtocounter{footnote}{-1}%
  \endgroup
}

\newcommand{\Real}[0]{\mathbb{R}}

\newcommand{\V}[0]{\mathbf{V}}

\newcommand{\p}[0]{\mathbf{p}}

\newcommand{\bK}[0]{\mathbf{K}}
\newcommand{\bH}[0]{\mathbf{H}}
\newcommand{\bR}[0]{\mathbf{R}}
\newcommand{\bX}[0]{\mathbf{X}}

\newcommand{\bZ}[0]{\mathbf{Z}}
\newcommand{\bz}[0]{\mathbf{z}}
\newcommand{\bt}[0]{\mathbf{t}}

\newcommand{\bM}[0]{\mathbf{M}}

\newcommand{\bs}[0]{\mathbf{s}}
\DeclareMathOperator{\Tr}{Tr}

\newif\ifdraft
\drafttrue

\definecolor{orange}{rgb}{1,0.5,0}

\ifdraft

    \newcommand{\LC}[1]{{\color{blue}{\textbf{LC: #1}}}}
    \newcommand{\MS}[1]{{\color{green}{\textbf{MS: #1}}}}
    \newcommand{\PF}[1]{{\color{red}{\textbf{PF: #1}}}}
    \newcommand{\PMN}[1]{{\color{orange}{\textbf{PMN: #1}}}}
    
\else

    \newcommand{\LC}[1]{}
    \newcommand{\MS}[1]{}
    \newcommand{\PF}[1]{}
    \newcommand{\PMN}[1]{}
\fi

%% file: 0_abstract.tex

\begin{abstract}

Given an image sequence featuring a portion of a sports field filmed by a moving and uncalibrated camera,
such as the one of a smartphone, our goal is to compute automatically and in real-time the focal length and
extrinsic camera parameters for each image in the sequence without using {\it a priori} knowledges of the
position and orientation of the camera. To this end, we propose a novel framework that combines accurate
localization and robust identification of specific keypoints in the image by using a fully-convolutional
deep architecture.
Our algorithm exploits both the field lines and the players' image locations, assuming their ground plane positions to be given,
to achieve accuracy and robustness that is beyond the current state of the art.
We will demonstrate its effectiveness on challenging soccer, basketball, and volleyball benchmark datasets.


\end{abstract}

%% file: 1_introduction.tex

\section{Introduction}

Accurate camera registration is a prerequisite for many applications such as augmented reality or 3D reconstruction.
It is now a commercial reality in well-textured environments and when additional sensors can be used to supplement the camera.
However, sports arenas such as the one depicted in Fig.~\ref{fig:intro} pose special challenges.
The presence of the well-marked lines helps, but they provide highly repetitive patterns and very little
texture. Furthermore, the players often occlude the landmarks that could be used for disambiguation. Finally, challenging lighting conditions are prevalent outdoors and not uncommon indoors, as shown in the figure.

\input{fig_intro}

As a result, traditional keypoint-based methods typically fail in such scenarios~\cite{Sattler18}.
An alternative is to explicitly use edge and line information~\cite{Homayounfar17,Sharma18, Jiang19}, but these methods tend to be slow,
to use sensitive parameterization based on vanishing points or to use prior knowledge of the position of the camera.
With the advent of deep learning, direct regression from the image to the camera pose has become an option~\cite{Kendall15a}, 
but it often fails to deliver accurate results.

In this paper, we propose a novel framework that combines accurate localization and robust identification
of specific keypoints in the image by using a fully-convolutional deep architecture~\cite{Ronneberger15}.
In practice, these keypoints are taken to be intersections between ground lines, and the network leverages
the fact that they do not overlap to drastically reduce inference time. These keypoints can then be used
to directly compute the homography from the image plane to the ground plane, along with the camera focal
length and extrinsic parameters, from single images. 
When using video sequences,
the unique identities of the keypoints make it easy to
impose temporal consistency and improve robustness.
Finally, when the ground location of the players is known, it can be used to further  improve accuracy and robustness of the estimations. To demonstrate this, we use a commercial system~\cite{SecondSpectrum} that uses a set of fixed cameras to compute these locations. This enables us to use the  players' feet as additional landmarks to increase robustness to narrow fields of views and lack of visible lines. 

We will show that our method outperforms the state-of-the-art methods for soccer scenarios~\cite{Homayounfar17,Sharma18, Jiang19},
which are the only ones for which there are published results. In addition, as publicly available datasets
in this subject are rare, we will introduce and demonstrate the effectiveness of our system in challenging basketball,
volleyball, and soccer scenarios that feature difficult light condition, motion blur, and narrow fields of view.
We will make the basketball and volleyball datasets publicly available.

In short, our contribution is a fast, robust, and generic framework that can handle much more challenging
situations than existing approaches. We leverage the fact that keypoints on a plane do not overlap
to drastically reduce inference time, thus enabling the detection of a high number of interest points.
In addition, we exploit the position of the players to further increase robustness and accuracy in images
lacking of visible features. Our method easily operates at 20-30 frames per second on a desktop computer with an Nvidia TITAN X (Pascal) GPU.


%% file: fig_intro.tex

\begin{figure}[t]
\centering
	\includegraphics[bb=0 0 1239 699, width=\columnwidth]{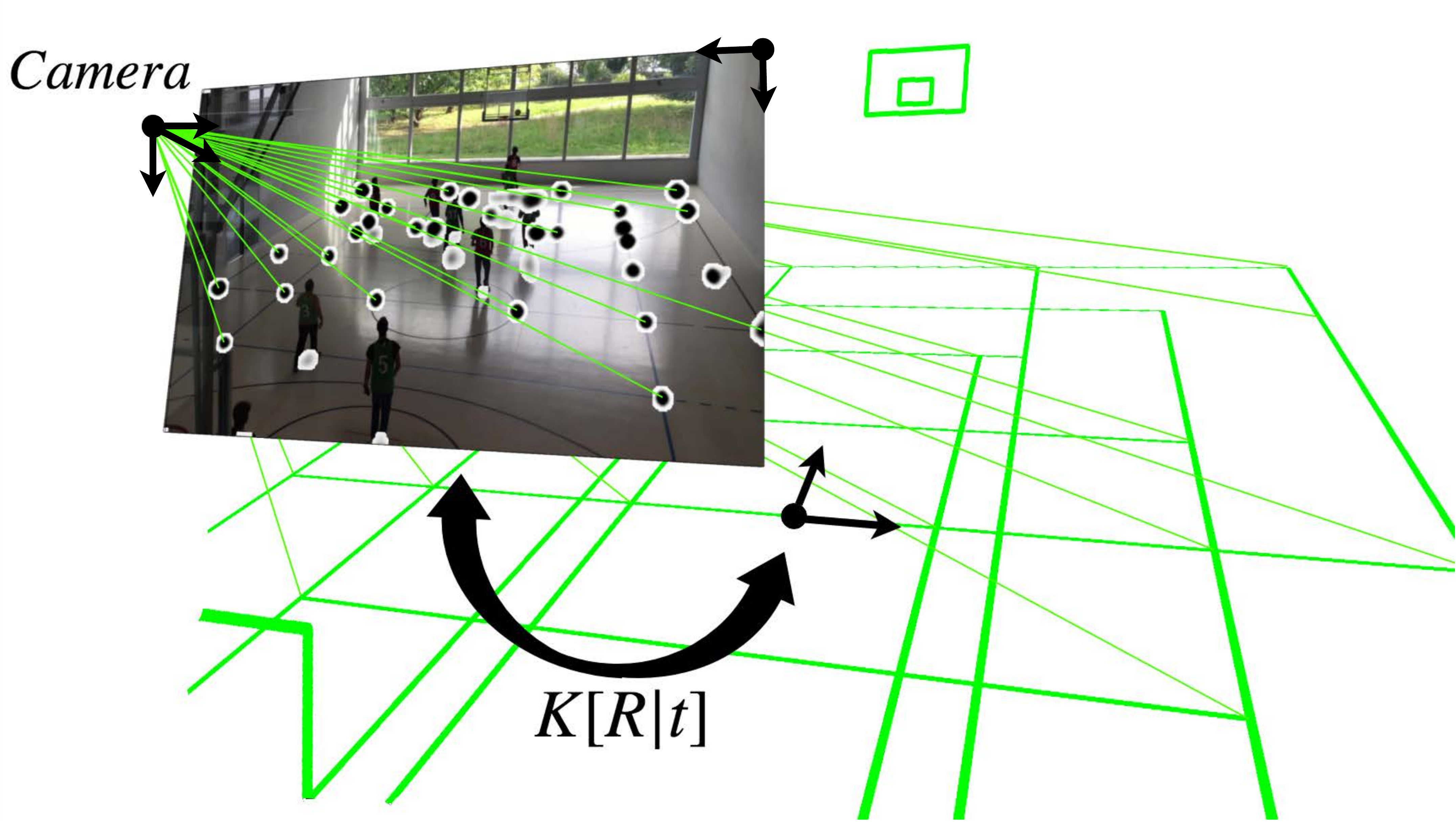} 
	\caption{\small {\bf Camera Pose Estimation from Correspondences.} Even though the lighting is bad, we can reliably detect preselected intersections of court lines. This yields 2D to 3D correspondences that we can use to compute the focal length and camera extrinsic parameters. If the position of the players is known, we can detect them and use them to instantiate additional 3D to 2D correspondences. \label{fig:intro}}
\end{figure}

%% file: 2_related_work.tex

\section{Related Work}
\label{sec:related}

As the dimensions of sports fields are known and 3D models are available, a naive approach to camera pose
estimation would be to look for projections of specific parts of the models in each image, establish 3D-to-2D
correspondences, and use them to compute the camera parameters.
Unfortunately, because the patterns in sports arenas and fields are repetitive, occlusions
frequent, or poor lighting, correspondences established using traditional methods such as SIFT~\cite{Lowe04}, SURF~\cite{Bay08},
or BRIEF~\cite{Calonder10b} are unreliable, which makes this approach prone to failure.
More specialized methods have therefore been proposed to overcome these difficulties by leveraging the specificities of sports fields without resorting to using additional sensors. 

In the case of soccer, the field is large and the lines delimiting it are widely separated.
As a result, in many camera views, too few are visible for reliable camera registration.
In~\cite{Chen18}, this problem is mitigated by using a two-point method.
The approach is effective but also very restrictive because it requires prior knowledge of the position and rotation axis of the camera.
In~\cite{Liu18} the lines of the field are used to compute a homography while in~\cite{Alvarez14} the mathematical characterization of the central circle is used to overcome the shortage of features.
Similarly, in~\cite{Gupta11}, points, lines, and ellipses are exploited to localize hockey rinks.
This can be effective for specific views but lacks generality.
In the results section, we will show that using the players' positions
is most beneficial when only few of the court lines are visible in the image.

In~\cite{Sharma18, Chen19} homography estimation relies on a dictionary of precomputed features of synthetic edge images and corresponding poses.
For a given input image, they first perform a nearest-neighbor search to find the most similar one in the database.
Then, the candidate homography is refined using image features.
Temporal consistency and smoothness is enforced in~\cite{Sharma18} over the estimates over time.
The limiting factor in these approaches is the variability of the potential poses.
Both methods use the fact that the camera is in a fixed position to reduce the size of the dictionary which would be very large otherwise.
In our approach, we have no constraints on the position and orientation of the camera and the variability of the poses does not affect the performance.

In~\cite{Homayounfar17} the problem is approached differently. Estimating the homography relating the image plane to a soccer
field is treated as a branch and bound inference in a Markov random field (MRF) whose energy is minimized when the
image and a generative model agree. The image is first segmented using a deep network to find lines, circles, and
grassy areas, and finally vanishing points. The vanishing points are used to constrain the search for the homography matrix and speed up the energy minimization. The limiting factor in this method is in the estimation of the vanishing points which involves a computation known to be error-prone when the perspective distortion is severe. Our approach involves no such computation.

In~\cite{Jiang19}, a framework that minimizes an inferred registration error is proposed for accurate localization of the field. Two deep networks are used. The first regresses an estimate of the homography characterizing the field. The second, the registration error between the current frame and the model projected with the estimate. 
Through differentiation in the second network, the homography is refined to minimize the registration error. The error estimation process and the refinement are then repeated multiple times until convergence. 
This method has the potential to produce accurate poses; however, it requires a relatively good initial estimate from the first network to converge. In addition, at each iteration the model of the court needs to be warped and forward passed into the second network making this approach slow.

In the experiment section we will compare our results to those of a traditional descriptor-based approach~\cite{Lowe04}
along with a newer end-to-end regression network~\cite{Kendall15a} and the approaches of~\cite{Sharma18, Homayounfar17, Jiang19} as they have demonstrated good results. 

%% file: 3_method.tex

\section{Approach}
\label{sec:approach}

\input{fig_keypoints}

Given an image sequence featuring a portion of a sports field filmed by a moving and
uncalibrated camera, such as the one of a smartphone, our goal is to compute in
real-time the focal length and extrinsic camera parameters for each image in the
sequence without using {\it a priori} knowledge of the camera position and orientation.

Our approach relies on two information sources that are more dependable and almost always available in images of sports fields.
The primary one comprises the lines painted on the ground, their intersections, and the corners they define, such as those depicted
by red dots in Fig.~\ref{fig:keypoints}.
We assign a unique identity to each one of them and refer to them as {\it semantic keypoints}.
The secondary source and optional one come from the players. 
We detect the projection of their center of mass on the ground in the images using a multi-camera setup and refer to them as {\it player keypoints}.
Since exploiting the identities of the players would require difficult tracking of
the number on the jerseys from single-view, we treat the player keypoints as points that do not
have a specific identity. The player keypoints are represented by the blue crosses in Fig.~\ref{fig:keypoints}.

To overcome the issues related to difficult lighting condition and poorly textured scenes we rely on a
fully-convolutional U-Net architecture~\cite{Ronneberger15} that combines global and local information
for simultaneous description and accurate localization of both kinds of keypoints.
This architecture has proved very effective for image segmentation purposes, and we will show that it is just as effective for our purposes.
First, we use the semantic keypoints localized in the image to compute an initial estimate of the homography that
maps the image plane to the field. Then, we use the players keypoints to refine the estimation.
The homography is then decomposed into intrinsic and extrinsic parameters. Finally, we use a particle filter to enforce robustness over time.

Fig.~\ref{fig:approach} summarizes our approach. We now formalize it and describe its individual components in more detail.

\subsection{Formalization}
\label{sec:formal}

\input{fig_approach}

Let $\{I^t\}_{t=1}^T$~be a sequence of~$T$ images of a sport field taken with a possibly uncalibrated camera.
The camera pose at time~$t$ is represented by a $3\times 4$ transformation matrix
$\bM^t = \left[\mathbf{R}^t\mid\mathbf{t}^t\right]$, where  $\mathbf{R}$ is a rotation matrix and $\mathbf{t}^t$ a
translation vector. $\bM^t$ is parameterized by 6 extrinsic parameters.
Similarly, the camera internal calibration is given by a $3\times 3$ matrix $\bK^t$ parameterized by 5 intrinsic parameters.
Formally, our problem is to find the state vector~$\bX^t=[\bK^t,\bM^t]$ for each image of the sequence.

We assume complete knowledge about the position of field lines.
In particular, we know the world coordinates~$\bZ_S$ of
a set of semantic keypoints in the sport field that have been manually selected once and for all, such as those depicted
by red dots in Fig.~\ref{fig:keypoints}.
Optionally, we can also assume knowledge of the position of the players on the
sport field at each time step~$t$, that is, the world coordinates~$\bZ_P^t$ of the projection of their center of gravity
onto the ground plane.
In practice, the players' positions in world coordinates are computed using a multi-camera system~\cite{SecondSpectrum} assumed to be synchronized with the mobile camera. The estimated position can be slightly imprecise when the players jump. However, the resulting error is small enough to be neglected. The player keypoints are shown as blue crosses in Fig.~\ref{fig:keypoints}.

Given an image~$I^t$ of the sequence, our method estimates the 2D-image locations $\hat \bz_S^t$ of the semantic keypoints
and $\hat \bz_P^t$  of the player's center of gravity. We then match them to their known 3D locations ~$\bZ_S$ and~$\bZ_P^t$.
From the resulting 3D-to-2D correspondences we compute the homography~$\bH^t$ between ground plane and the image, which is
then further decomposed into $\bK^t$ and~$\bM^t$ as described in Appendix.

A typical basketball court or soccer field is roughly symmetric with respect to its long and short axis. As a result, it can be hard to distinguish views taken from one corner from those taken from the opposite one as the keypoints will look the same. However, if we know on what side of the field the camera is, this ambiguity disappears and we can assign a unique identity to all keypoints, which is what we do. More specifically, during training we swap the identities of the symmetric points when the camera moves to the other side of the court. By doing so we maintain the same identity to similar feature points and provide coherent inputs to our network.
If the court has logos that are {\it not} symmetric, two networks have to be trained, one for each side of the playing field. 

\subsection{Detecting Keypoints}
\label{sec:keypoints}

\input{fig_detect}

To detect the keypoints and compute their 2D locations in individual images, we train a single U-Net deep
network~\cite{Ronneberger15} that jointly locates semantic and player keypoints in the image.
We configure the network to perform pixel-wise classification with standard multi-class cross-entropy loss.
The network produces a volume that encodes the identity and presence of a keypoint so that the pixels that
are within a distance $b$ from it will be assigned the class corresponding to it.

\parag{Output Volume.}
We dimension our network to take as input an RGB image $I \in \Real^{H \times W \times 3}$ and return a
volume $\V = \{\V_0, ..., \V_{J+1}\}$ composed of $J+2$ channels where $J$ is the number of semantic keypoints. These
have a unique identity, therefore, channels $\V_j$ with $j$ in the range $\{1,...,J\}$ are used to encode their locations. On the other hand, the player keypoints are all assigned the same identity, to this end,
we define a single channel $\V_{J+1}$ to encode their locations. To assign a class also to locations where there is
no keypoint we take $\V_0$ to be the background channel.

Let $\bz_{S|j}$ be the projected $j$-th semantic keypoint and $\V^*_{j}$ its associated ground-truth channel,
we set to $1$ the pixels that are at a distance $b$ from the interest point and $0$ elsewhere. In the same manner,
we set the pixels of $\V^*_{J+1}$ to $1$ at a location where there is a player keypoint and $0$ elsewhere.
If two keypoints are within $2\cdot b$ of each other, the pixels are assigned the class of the closest one.
Finally, we set the background channel $\V^*_0$ so that for any value at location $\p \in \Real^2$ it satisfies $\sum_i \V_i(\p) = 1$.

\parag{Training.}
The output of our network is a volume $\V \in \lbrack 0,1 \rbrack^{H\times W\times (J+2)}$
that encodes class probabilities for each pixel in the image, that is the probability of a pixel belonging to one of
the $J+2$ classes defining the keypoints' identities and the background. This is achieved using a pixel-wise softmax layer.
During training, the ground-truth keypoints' locations $\bZ_S$ and $\bZ_P^t$ are projected for a given image $I^t$ using
the associated ground-truth homography. Then, these projections are used to create the ground-truth output
volume $\V^* \in \{0,1\}^{H\times W\times (J+2)}$ as described in the previous paragraph.
In addition to volume $\V^*$, we create weights to compensate class imbalance.
These are defined for a given class as the fraction of the total number of pixels in the image and the number
of pixels that belong to that class.

As will be discussed in Section~\ref{sec:datasets}, our training data comprise sequences of varying lengths taken
from different viewpoints with annotations for both the semantic keypoints and the players' locations.
At every training iteration, we choose our minibatches by taking into consideration the frequency of a viewpoint.
In other words, images from short sequences are more likely to be chosen than images from long ones.
This tends to make the distribution of viewpoints more even.

Finally, to increase global context awareness, we use an augmentation method named SpatialDropout~\cite{Tompson15} during training to force the
network to use the information surrounding a keypoint to infer its position.
At every training iteration, we randomly create boxes of different sizes and zero-out the pixels of the input image that are within them.
The number of boxes so as their sizes and positions are drawn from uniform distributions. As a result and as shown in Fig.~\ref{fig:detect}, keypoints
can be correctly detected and localized even when they are occluded by a player.

\parag{Inference.}
At run-time, we leverage the fact that keypoints defined on a plane do not overlap in projection to drastically reduce inference time.
The background channel $\V_0$ encodes all the information required to locate a keypoint; therefore, we perform non-minimum suppression
on this channel only and then assess their identities by looking for the index of the maximum in the corresponding column in the volume. This enables us to handle  many interest points in real-time.

\subsection{Estimating Intrinsic and Extrinsic Parameters}
\label{sec:estimation}

Having detected semantic keypoint locations $\hat \bz_S^t$, that is, markings on the ground, and player keypoints $\hat \bz_P^t$, that is,
the projection of player's center of gravity in image $I_t$, we can now exploit them to recover the camera parameters.
Since the camera focal length is not known {\it a priori}, the camera extrinsics cannot be computed directly.
To this end, we first compute a homography $\bH^t$ from the image plane to the field then we estimate intrinsics and
the extrinsic parameters $\bK^t$ and $\bM^t$ from $\bH^t$ as described in the Appendix.

As discussed at the end of Section~\ref{sec:formal}, the locations of the semantic keypoints can be readily used to estimate a homography because they are assigned unique identities
that translate into a 3D-to-2D correspondence between a 3D point on the playing field and a 2D image location.

By contrast, exploiting the player keypoints that can be the projection of one of many 3D locations requires establishing the correspondences.
Doing so by brute force search would be computationally infeasible in real-time, and even using more sophisticated methods
that leverage {\it a priori} knowledge about the camera position~\cite{David04,Moreno08} can be slow.
Instead, we use a simple yet effective two-step approach: Given image $I^t$, we use the semantic keypoints to compute
a first estimate of the homography $\bH_0^t$. This allows us to back-project the detected player locations $\hat \bz_P^t$
from the image plane to world coordinates and associate the back-projected points to
the closest ground-truth positions $\bZ_P^t$.
Finally, we use the newly defined players' correspondences with the already known semantic ones to estimate a new homography $\bH_1^t$.

This approach enables us to use players data to produce a more accurate mapping that translates to better estimates of the focal length and the pose.

\subsection{Enforcing Temporal Consistency}
\label{sec:temporal}

Using the approach described above, keypoints can be found independently in individual images and used to compute a
homography and derive camera parameters for each. However, in a video sequence acquired with a moving camera, this
fails to exploit the fact that its motion may be shaky but is not arbitrary. To do so, we rely on a particle filtering
approach known as {\it condensation}~\cite{Isard98b} to enforce temporal consistency on the pose $\bM^t$, with the
intrinsics $\bK^t$ being updated at each iteration to allow the focal length to change.

The idea underlying the condensation algorithm is to numerically approximate and refine the probability density
function $p(\bM^t |\hat \bz^t)$ over time. A set of $N$ random poses called particles $\bs^t_n$ with associated
weights $\pi^t_n$ approximate the posterior distribution
\begin{equation}
\hat p(\bM^t |\hat \bz^t) = \sum_{n=1}^{N} \pi_{n}^t \delta(\bM^t - \bs^t_{n}),
\label{eq:posterior}
\end{equation}
where $\delta(.)$ is the Dirac delta function. At every iteration, the particles are generated, transformed and
also discarded based on their weights $\pi^t_n$. These are at the heart of this procedure.
The larger they are for a given particle, the more likely it is to be retained. They should therefore be chosen
so that particles that correspond to likely extrinsic parameters, that is, parameters that yield low re-projection
errors are assigned a high weight.

To this end, we use the extrinsic parameters associated to the particles to project the ground-truth 3D points
and compute the mean error distance from the projections to the estimated positions $\hat \bz^t$. For the
semantic keypoints, we compute the distance $\xi^t_{S|n}$ to the corresponding predicted 2D location.
For the player ones whose identity is unknown, we search for the detection closest to the projection and use
it to compute the error $\xi^t_{P|n}$. We assume a Gaussian model for both error components $\xi^t_{S|n}$ and $\xi^t_{P|n}$. Therefore, we take the weight of particle $n$-th to be

\begin{equation}
\pi_n^t = \alpha\, \mathrm{exp} \Big[\Big(\frac{-\xi^t_{S|n}}{\sqrt{2}\sigma_{S}}\Big)^2\Big] + (1-\alpha)\,\mathrm{exp}\Big[\Big(\frac{-\xi^t_{P|n}}{\sqrt{2}\sigma_{P}}\Big)^2\Big],
\label{eq:weights}
\end{equation}
where $\sigma_{S}$ and $\sigma_{G}$ control the importance of a particle based on its error, $\alpha$ instead balances
the two contributions.
Intuitively, if the error for a given particle is close to zero the associated weight will be close to one.
The new state is then taken to be the expected value of the posterior
$E[p(\bM^t|\hat \bz^t)] \approx \sum_{n=1}^{N} \pi_n^t \bs_n^t$. We describe the whole framework procedure in more
detail in Appendix~\ref{sec:framework}.

%% file: fig_keypoints.tex

\begin{figure}[]
	\centering
	\includegraphics[bb=0.00 0.00 841.89 490, width=\linewidth]{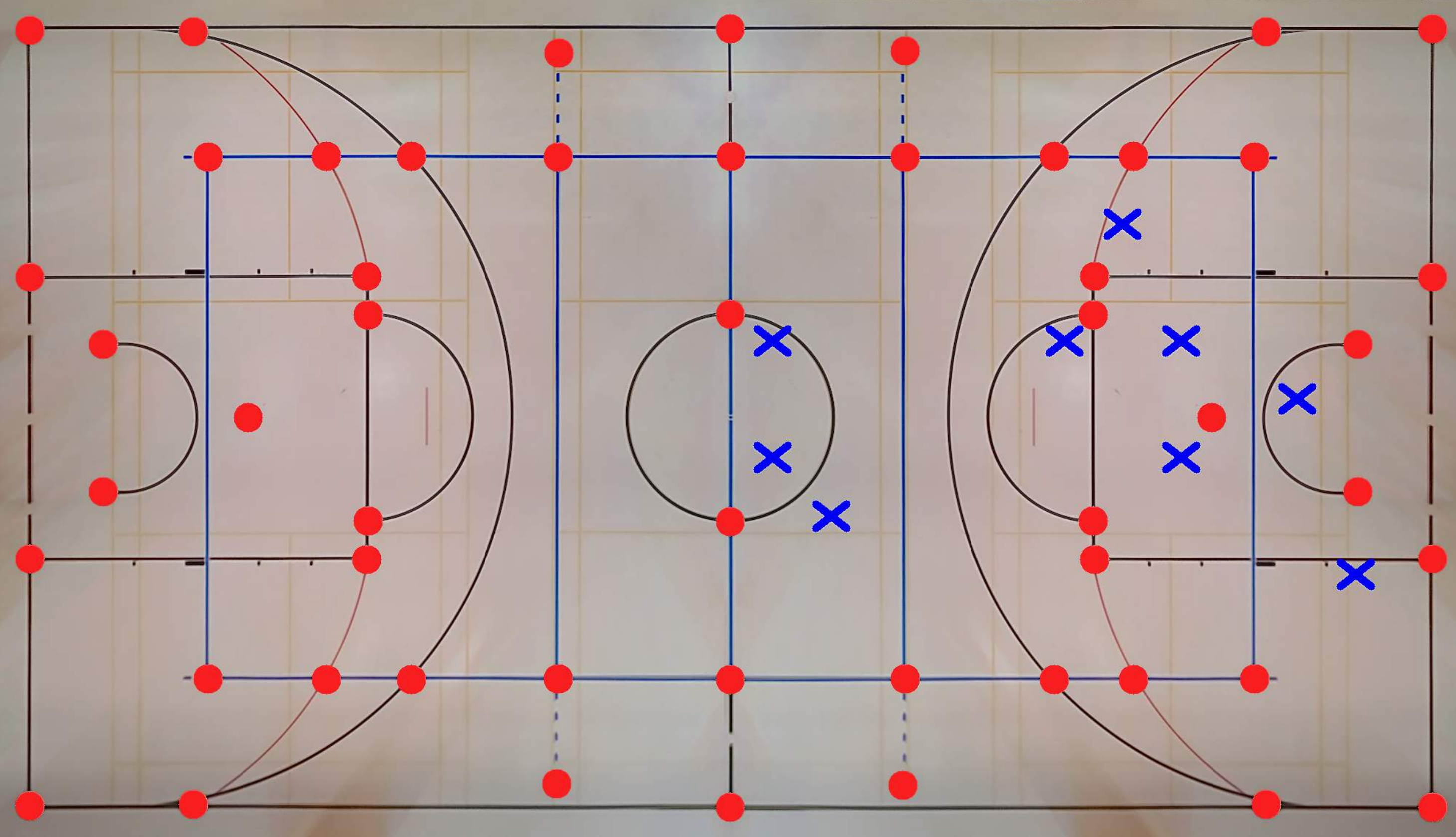}
	\vspace{-3pt}
	\vspace{-2mm}
	\caption{\textbf{Semantic and Player Keypoints.} The red dots denote semantic keypoints.
	The blue crosses represent players' locations that become our player keypoints when they are available.
	\label{fig:keypoints}}
\end{figure}

%% file: fig_approach.tex

\begin{figure}
	\centering
	\includegraphics[width=\columnwidth]{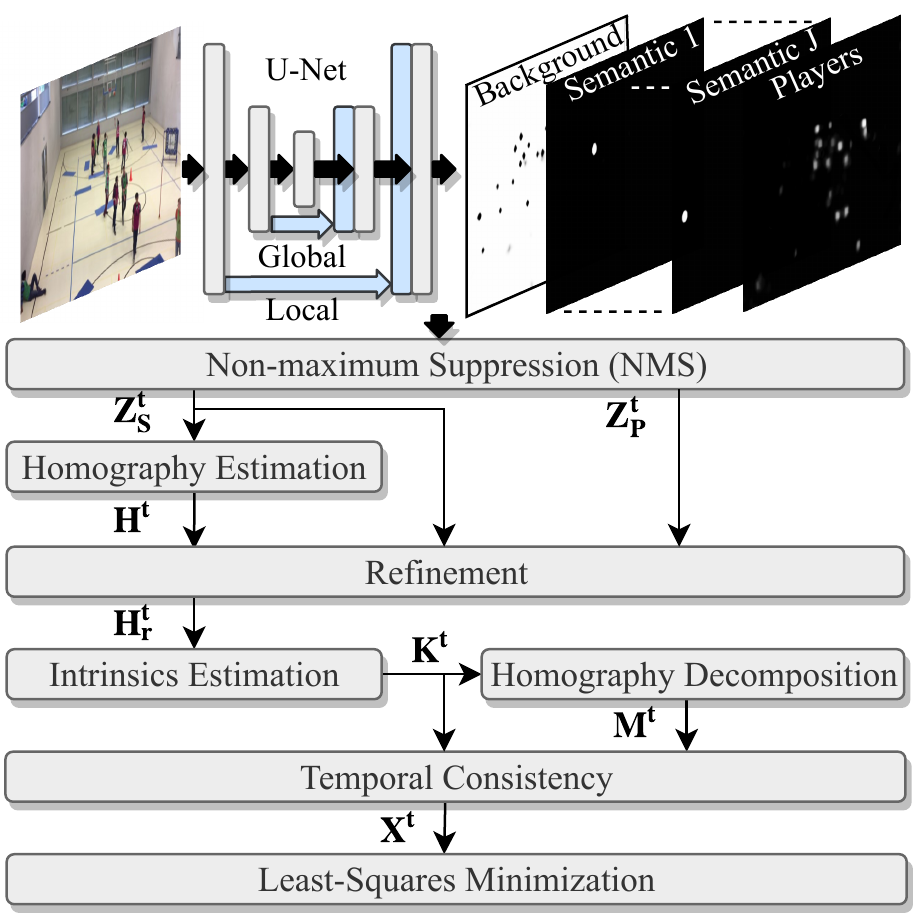}
	\vspace{-2mm}
	\caption{\small {\bf Our approach.} We use a U-Net to detect the semantic and player keypoints. They are then used to compute the homographies from the ground to the image planes in individual images. Finally, the camera parameters are inferred from these homographies and refined by imposing temporal consistency and non-linear least-square minimization. \label{fig:approach}}
\end{figure}

%% file: fig_detect.tex

\begin{figure}[t]
	\centering
	\includegraphics[width=\linewidth]{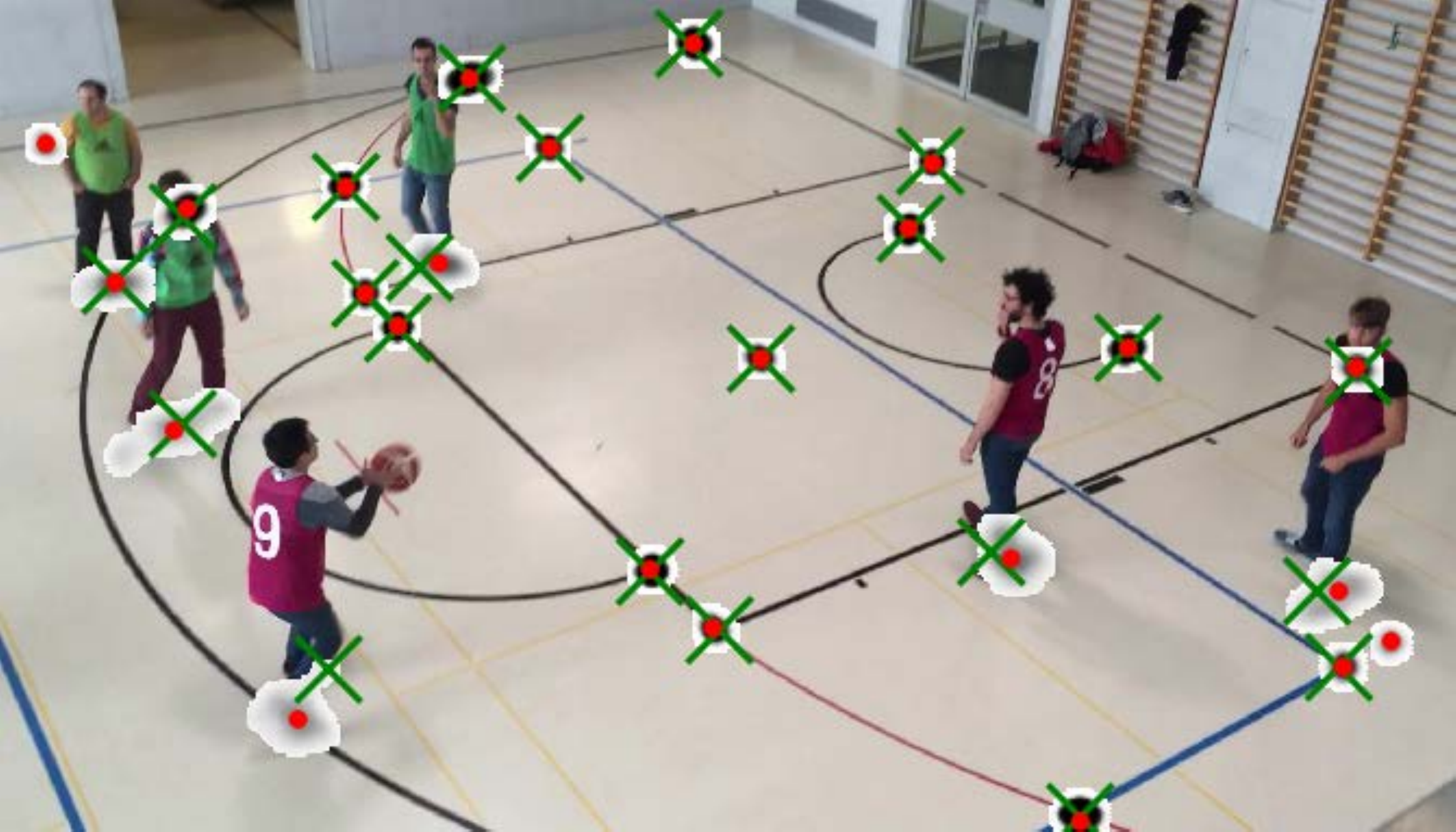}
	\vspace{-3mm}
	\caption{{\bf Detecting keypoints.} Example image with superimposed prediction.
	The crosses indicate the true projection of the interest points and the red dots
	the 2D locations found by our network. The gray/white patches represent the score
	returned by the U-Net. Even though some interest points are occluded, the network
	localizes them accurately. \label{fig:detect}}
\end{figure}

%% file: 4_experiments.tex

\newcommand{\basket}[0]{Basketball}
\newcommand{\volley}[0]{Volleyball}
\newcommand{\soccerM}[0]{Soccer MLS}
\newcommand{\soccerW}[0]{Soccer World Cup}

\newcommand{\bbasket}[0]{\textbf{Basketball}}
\newcommand{\bvolley}[0]{\textbf{Volleyball}}
\newcommand{\bsoccerM}[0]{\textbf{Soccer MLS}}
\newcommand{\bsoccerW}[0]{\textbf{Soccer World Cup}}

\newcommand{\sift}[0]{{SIFT}}
\newcommand{\siftK}[0]{{Keyframe SIFT}}
\newcommand{\siftT}[0]{{Top View  SIFT}}
\newcommand{\poseN}[0]{{PoseNet}}
\newcommand{\bbmrf}[0]{{Branch and Bound}}
\newcommand{\hogdict}[0]{{Synthetic Dictionary}}
\newcommand{\errorref}[0]{{Learned Errors}}

\newcommand{\bsift}[0]{\textbf{SIFT}}
\newcommand{\bsiftK}[0]{\textbf{Keyframe SIFT}}
\newcommand{\bsiftT}[0]{\textbf{Top View  SIFT}}
\newcommand{\bposeN}[0]{\textbf{PoseNet}}
\newcommand{\bbbmrf}[0]{\textbf{Branch and Bound}}
\newcommand{\bhogdict}[0]{\textbf{Synthetic dictionary}}
\newcommand{\berrorref}[0]{\textbf{Learned Errors}}

\newcommand{\ours}[0]{{OURS}}
\newcommand{\oursNOpf}[0]{{OURS w/o P.Filter}}
\newcommand{\oursNOsd}[0]{{OURS w/o S.Dropout}}
\newcommand{\oursNOpl}[0]{{OURS w/o Players}}

\newcommand{\bours}[0]{{\bf OURS}}
\newcommand{\boursNOpf}[0]{{\bf OURS w/o P.Filter}}
\newcommand{\boursNOsd}[0]{{\bf OURS w/o S.Dropout}}
\newcommand{\boursNOpl}[0]{{\bf OURS w/o Players}}

\section{Experiments}

\subsection{Datasets, Metrics, and Baselines}
\label{sec:datasets}

In this section we introduce the datasets we tested our method on, the metric we used to assess performance, and the baselines against which we compared ourselves.

\parag{Datasets.}
We tested our approach on the following datasets
\begin{itemize}

\item \bbasket{}.
We filmed an amateur basketball match on our campus using smartphones and moving around the field.
At the same time, 8 fixed and calibrated cameras were filming it, and we used their output to estimate the players' positions on ground.
The sequences contain a variable number of people ranging from 10 to 13 that are either running, walking or standing.
For each smartphone image, we estimated ground-truth poses using a semi-automated tool. It tracks interest points from image to image under the supervision of an operator. When the system looses track, the operator can click on points of interest and restart the process. In practice, this is much faster than doing everything manually. In this manner, it took us about 60 hours to compute homographies for all of the 50127 images forming 28 distinct sequences, 
some of which feature difficult light conditions and foreign objects such as gymnastic mats and other pieces of equipment occluding parts of the field. Manually annotating every frame would have taken at least six weeks. We used 12 sequences for training and 16 for testing. This dataset will be made publicly available.

\item \bvolley{}.
These volleyball sequences were filmed using broadcast cameras and are publicly available~\cite{Ibrahim16},
along with the corresponding players' positions. We again used the semi-automated tool described above to compute ground-truth
poses and intrinsic parameters that change over time in 12987 images coming from four different matches
and will also make them publicly available.
The images include players, referees and coaches but only the players, six in each team, were tracked.
We used two sequences for training and two for testing.

\item \bsoccerM{}.
We filmed a Major League soccer match using one moving smartphone and 10 fixed cameras to estimate the position of the players,
as for the \basket{} dataset. All the players and the three referees were tracked in this dataset for a total of 25 people. The focal length is constant between frames but different for each sequence.
We then used our semi-automated tool to compute ground-truth poses for 14160 images divided into 20 sequences from different
locations around the court.
We used 10 sequences for training and 10 for testing.

\item \bsoccerW{}.
This is a publicly-available dataset used in~\cite{Homayounfar17, Sharma18}. It comprises 395 images acquired by the broadcast cameras at the 2014 World Cup in Brazil.
The images have an associated homography and they are not in sequence.
Since the players' positions are not provided, we extracted them manually in every image in order to demonstrate their usefulness. We extracted the position of all the players and the referees that are visible in the images. 

\end{itemize}

\parag{Evaluation Metrics.}
We use five metrics to evaluate the recovered camera parameters:
intersection over union (IoU), reprojection error, angular error in degrees, translation error in meters and relative focal length error.
We report the mean, the median and the area under the curve (AUC) for each of them.

To compute the reprojection error we first project a grid of points defining the playing
surface onto the image, and then average their distances from their true locations.
Only the visible points in the image are taken into account.
For independence from the image size we normalize the resulting values by the image height.

The IoU is taken to be the shared area between the ground-truth model of the
court and the re-projected one divided by the area of the union of these two areas.
The IoU is one if the two coincide exactly and zero if they do not overlap at all.
In this work and in~\cite{Homayounfar17}, the entire sports field template is considered.
In~\cite{Sharma18}, the IoU is computed using only the area of the court that is visible in the image. 
We discourage the use of this version as it has an important flaw. 
Since this metric simply compares areas, cropping both ground-truth and model means removing the parts that are not aligned. These are in fact the ones that contribute negatively to the score. It is therefore easier to obtain perfect scores even if the estimate is far from correct. We give more explanations in Appendix~\ref{sec:iou_issue}.

We compute the angular error as $\arccos{[(\Tr(\bR_{gt}^T\cdot \bR_{est}) - 1)/2]}$ while the translation error as $||\bt_{gt}-\bt_{est}||_2$. The relative focal length error is defined as $|f_{gt}-f_{est}|/f_{gt}$.
Finally, the AUC is computed by sorting the errors in ascending order and by considering the values lower than a threshold. For IoU we take a threshold of $1$, for the reprojection error $0.1$, for the angular error $10^{\circ }$, $2.5$ m for the translation error and $0.1$ for the relative focal length error. 

\parag{Baselines and Variants.}
\label{sec:baselines}
We compare our method against the following approaches.
\begin{itemize}

\item \bsift{}~\cite{Lowe04}:
We use the OpenCV implementation of SIFT to locate and match interest points between an image and a set of reference images.
We manually select reference images from the training set in such a way to cover all viewpoints.
Given a query image, we attempt to match it against each reference image in turn.
We use the two-nearest-neighbor heuristic~\cite{Lowe04} with a distance ratio of 0.8 to reject
keypoints without a reliable correspondence.
The reference image that features the largest number of correspondences is used in conjunction
with RANSAC \cite{Fischler81} to compute a homography.

\item \bposeN{}~\cite{Kendall15a}: Direct regression from the query image to a translation and quaternion vectors.
We use Resnet-50~\cite{He16} pretrained on ImageNet. We replace the last average pooling and fully-connected layers
with a ReLU activation followed by a $1\times 1$ convolutional layer to reduce the number of features of the
activation volume from $2048$ to $512$ then a fully-connected layer to output the 7 elements vector.
Instead of feeding an image we feed the accumulator of the Hough transform, this produces better results overall.
In addition, we normalize the values of translation and quaternions vectors using the mean of the training distribution.
Finally, we set the balance parameter $\beta$ of the loss to $1e-3$ and train the network for $50000$ iterations
with Adam and learning rate set to $1e-4$.

\item  \bbbmrf {}~\cite{Homayounfar17}: A branch and bound approach using lines and circles as cues to estimate a homography.

\item \bhogdict{}~\cite{Sharma18}: Nearest-neighbor search over a precomputed dictionary of synthetic edge images.

\item \berrorref{}~\cite{Jiang19}: Homography refinement through iterative minimization of an inferred registration error.

\end{itemize}
The last three approaches are presented in more detail in Section~\ref{sec:related}. We also compare against the following variants of our own approach.
\begin{itemize}
\item \bours{}: Our complete method using semantic keypoints, player positions, refinement stages, spatial-dropout and particle filter.
\item \boursNOpl{}: Our method without using the players.
\item \boursNOpf{}: Our method without the particle filter, that is, without temporal consistency.
\item \boursNOsd{}: Our method without SpatialDropout \cite{Tompson15} for the increase in global context awareness.

\end{itemize}

\input{fig_sift_ours}
\input{fig_good_cases}
\input{fig_failure_cases}

\subsection{Implementation Details}

For all our experiments we train a U-Net architecture~\cite{Ronneberger15} from scratch using SpatialDropout~\cite{Tompson15}, pixelwise softmax as last layer, cross-entropy loss and ReLU activation.
For \basket{}, \volley{} and \soccerM{} the number of downsampling steps of the network are 4 whereas the number of filters
in the first layer to 32, in \soccerW{} to 5 and 48 respectively. We optimize the parameters of the network using
Adam~\cite{Kingma15} with learning rate set to $1e-4$. We resize the images by maintaining the original aspect ratio.
For \basket{} and \volley{} the height of the input images are 256 pixel, in \soccerM{} 360 pixels whereas in \soccerW{} 400 pixels.
At every iteration a patch of size $224\times 224$ is randomly cropped from the image and fed into the network. We use batch size of 4.
In non-maximum suppression, a local-maximum is considered a keypoint if its response is higher than 0.25, we discard the rest.
For all the experiments we use 300 particles for the filter with $\sigma_S$ and $\sigma_G$ set both to $2$, $\alpha=0.5$.
We run all the experiments on an Intel Xeon CPU E5-2680 2.50GHz and Nvidia TITAN X (Pascal) GPU.

\subsection{Qualitative and Quantitative Results}

\input{fig_plots}
\input{table_results_ours}
\input{table_results_wc}

Fig.~\ref{fig:siftVsOurs} demonstrates the effectiveness of our method in establishing correspondences.
Our method outperforms \bsift{} by a large margin. Fig.~\ref{fig:good_cases} depicts some qualitative
results for basketball, soccer and volleyball. The registration in these cases is accurate to the point
where the projected models' lines match almost perfectly the lines of the court.
By contrast, we present some failure cases in Fig.~\ref{fig:failure_cases}.
They are usually caused by the lack of visible lines, clutter, and narrow field of view.
We now turn to quantifying these successes and failures.

\parag{Comparison to the Baselines.}
We report comparative results on our benchmark datasets in Fig.~\ref{fig:plots} and Tab.~\ref{tab:results_ours}. 
The results for \bsoccerW{} are shown in Tab.~\ref{tab:results_wc}. 
As the \bsoccerW{} dataset does not include intrinsic and extrinsic parameters, we report IoU and reprojection error only. 
The code for \bhogdict{} and \berrorref{} are not publicly available, therefore,  we report the published results for \bsoccerW{}.
Note that in Tab.~\ref{tab:results_ours} we do not report the relative focal length errors for \bposeN{} as this method only produces rotation and translation matrices. 
To compute the other metrics for \bposeN{} we used the ground-truth intrinsics.

\bours{} does best overall with \boursNOpl{} a close second. Interestingly, \boursNOpl{} does slightly
better than  \bours{} on \bvolley{}. We take this to mean that, in this case, the keypoints are dense
enough for a precise estimate. Therefore using the players whose location cannot be detected very accurately does not help.
By contrast, for \bsoccerW{} and \bsoccerM{} where the field is larger and the keypoints fewer, using the players
is key to top performance. As shown by the median in Tab.~\ref{tab:results_ours} and by Fig.~\ref{fig:plots}, our method is extremely precise.
For most of the basketball and volleyball images, the reprojection error on our method is less
than 5 pixels on a full-hd image ($1080\times1920$), in soccer less than 7 pixels.
The translation error is less than 20 cm for most images of our basketball dataset, about 50 cm for volleyball and less the 3 m for soccer.

\parag{Ablation Study.}

\input{table_keypoints_configs_table}
\input{fig_keypoints_configs}

In Tab.~\ref{tab:results_ours}, we also report performance numbers for \boursNOpf{} and \boursNOsd{}.
They are consistently worse than those of our full approach, thus confirming the importance of the particle filter and of dropout.

A question that arises in practice is where to position the keypoints.
As the network uses both local and global information, keypoints can be placed anywhere on the field, however,
their position affects how precise is their localization in the image. Fig.~\ref{fig:keypointsConfigs} depicts three potential configurations.
We trained and tested our network using each one in turn. In Table~\ref{tab:keypointsConfigs}, we report the corresponding
average distance between projected ground-truth points and the detections, along with the proportion of inliers.
The average distance is computed using the detections that are within 5 pixels to the closest corresponding ground-truth point.

The two configurations with keypoints located at  corners and line intersections are more precise and perform very similarly.
The third configuration with regularly spaced keypoints that do not match any specific image feature does less well but
still yields a reasonable precision. This confirms our network's ability to account for context around the keypoints.

%% file: fig_sift_ours.tex

\begin{figure*}[]
\centering
	\includegraphics[width=\linewidth]{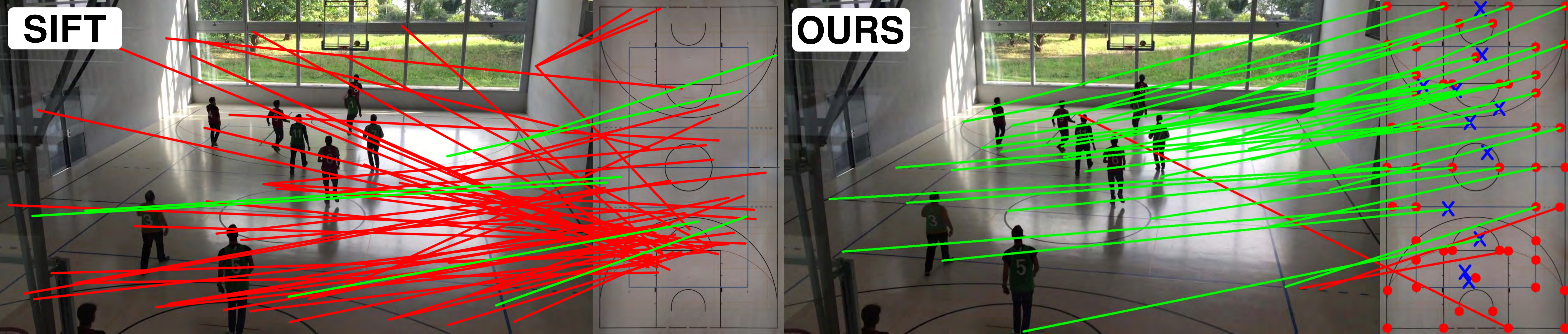}
	\vspace{-4mm}
	\caption{\small {\bf Robust Keypoints Detection.}
	{\bf (left)}   Putative correspondences drawn from the image to the model of the court
	using SIFT approach described in Section~\ref{sec:baselines} and {\bf (right)} by our method where red and blue dots are
	semantic and generic keypoints respectively.  
	Incorrect correspondences are shown as red lines, green otherwise.
	Even though the lighting is poor, our method (right) can reliably establish many correct correspondences,
	whereas those found using SIFT~\cite{Lowe04} (left) are mostly wrong.
	The other baselines methods do not appear in this figure as they are not keypoint based.
	\label{fig:siftVsOurs}}
\end{figure*}

%% file: fig_good_cases.tex

\begin{figure*}[h!]
  \centering
  \setlength{\tabcolsep}{1pt} 
  \renewcommand{\arraystretch}{0.2} 
  \resizebox{\linewidth}{!}{%
	  \begin{tabular}{@{}ccc@{}}
	    \includegraphics[width=.32\linewidth]{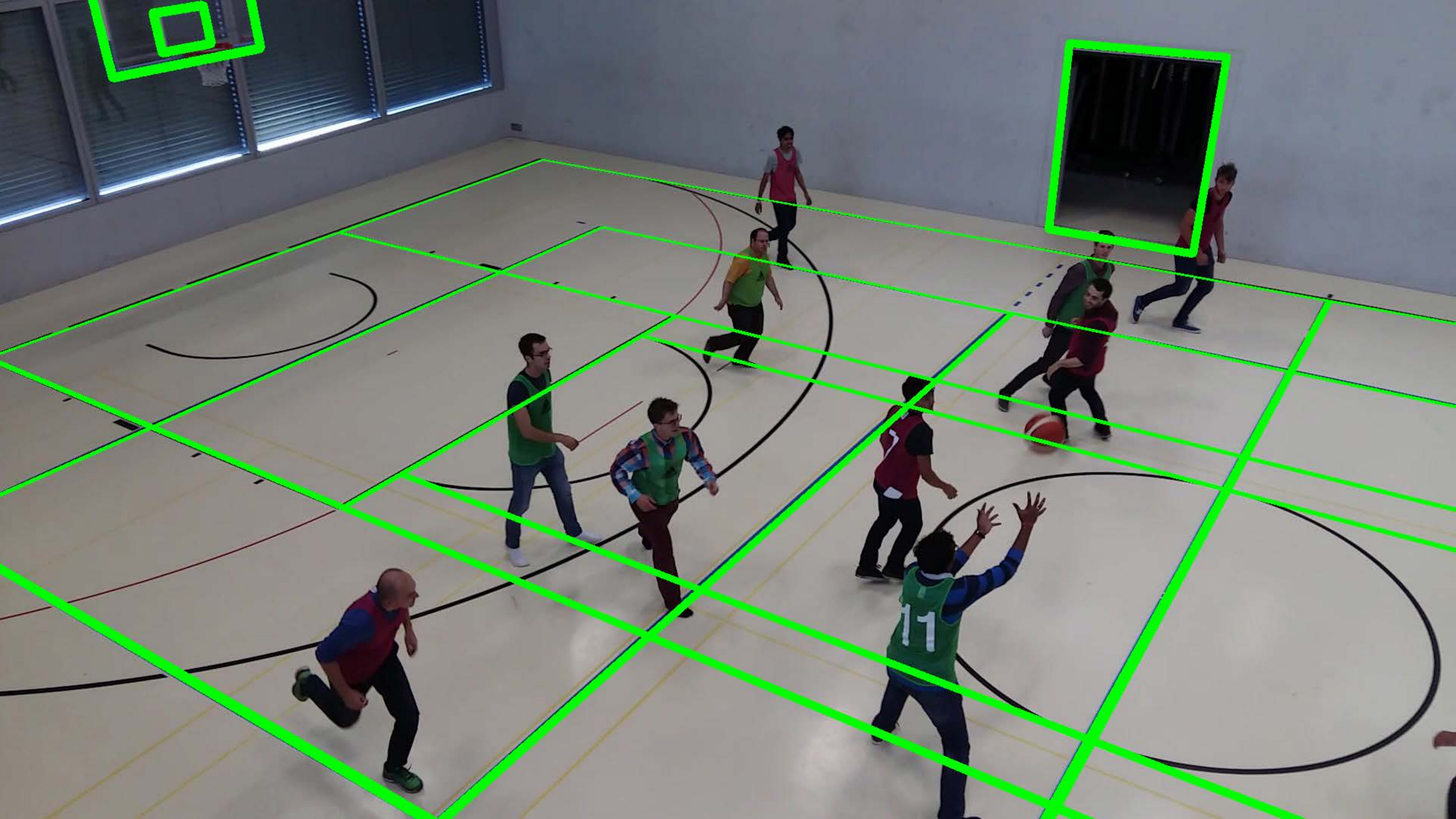} &  
	    \includegraphics[width=.32\textwidth]{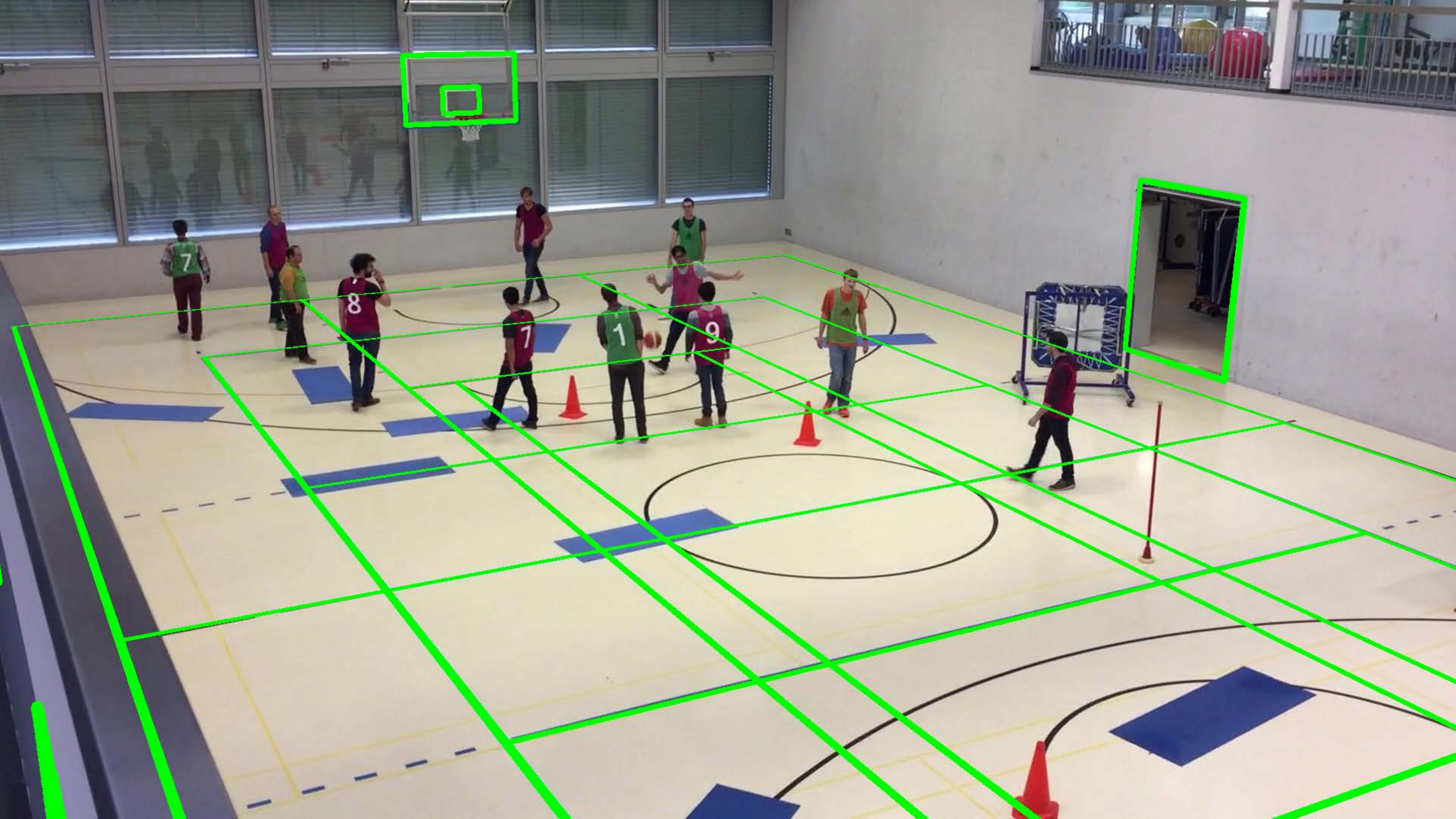} &  
	    \includegraphics[width=.32\textwidth]{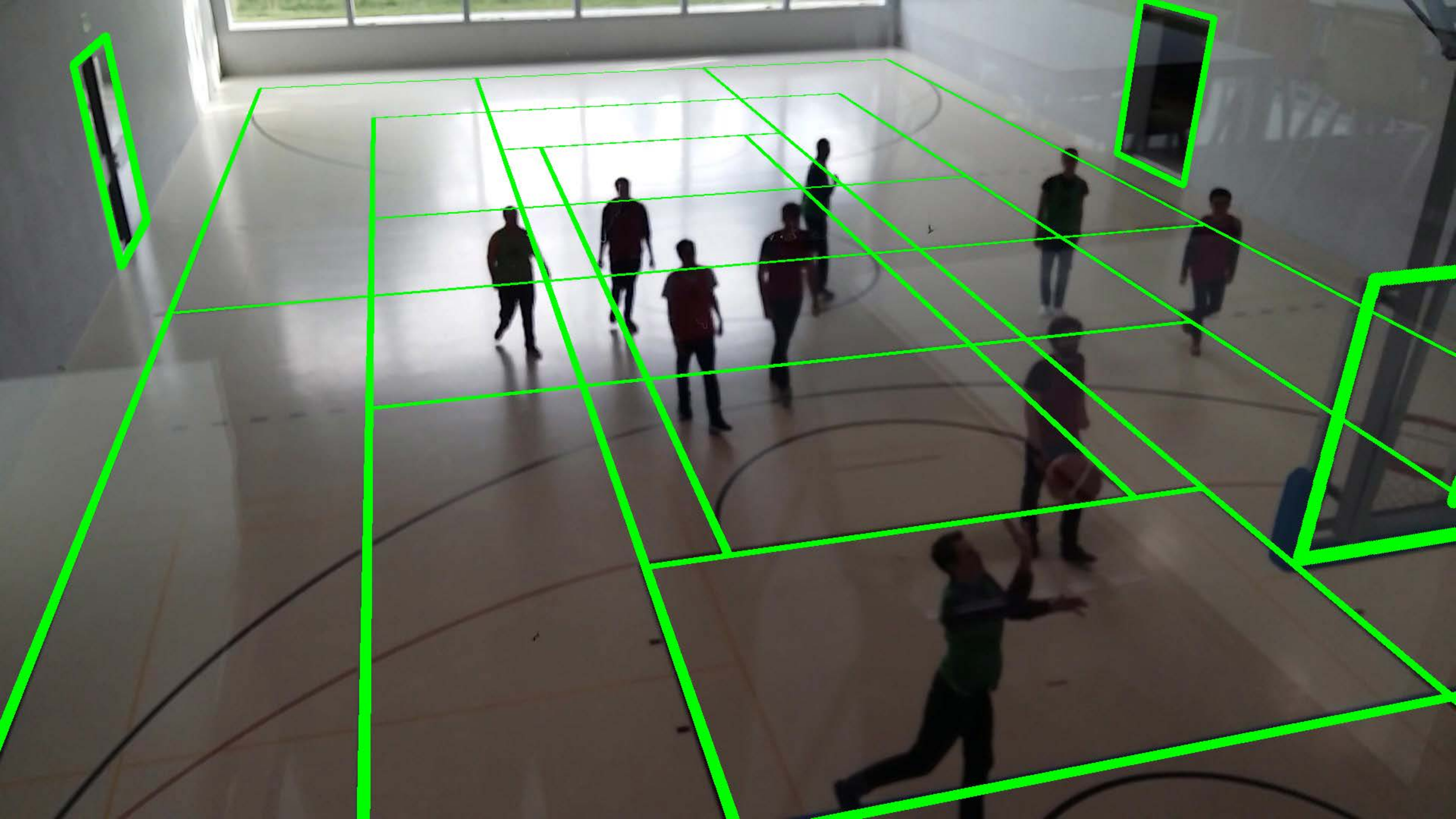} \\
	    \includegraphics[width=.32\textwidth]{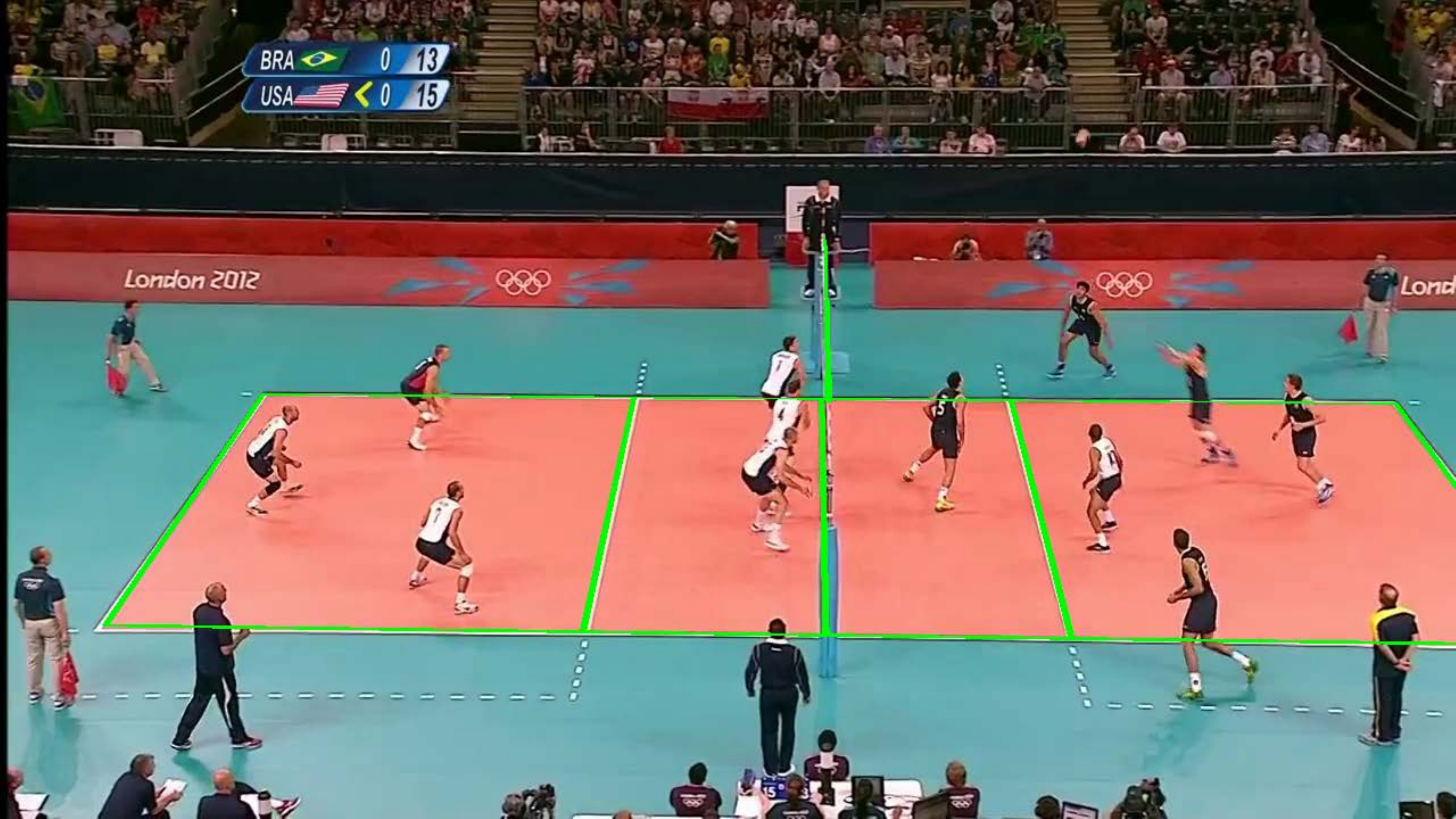} &
	    \includegraphics[width=.32\textwidth]{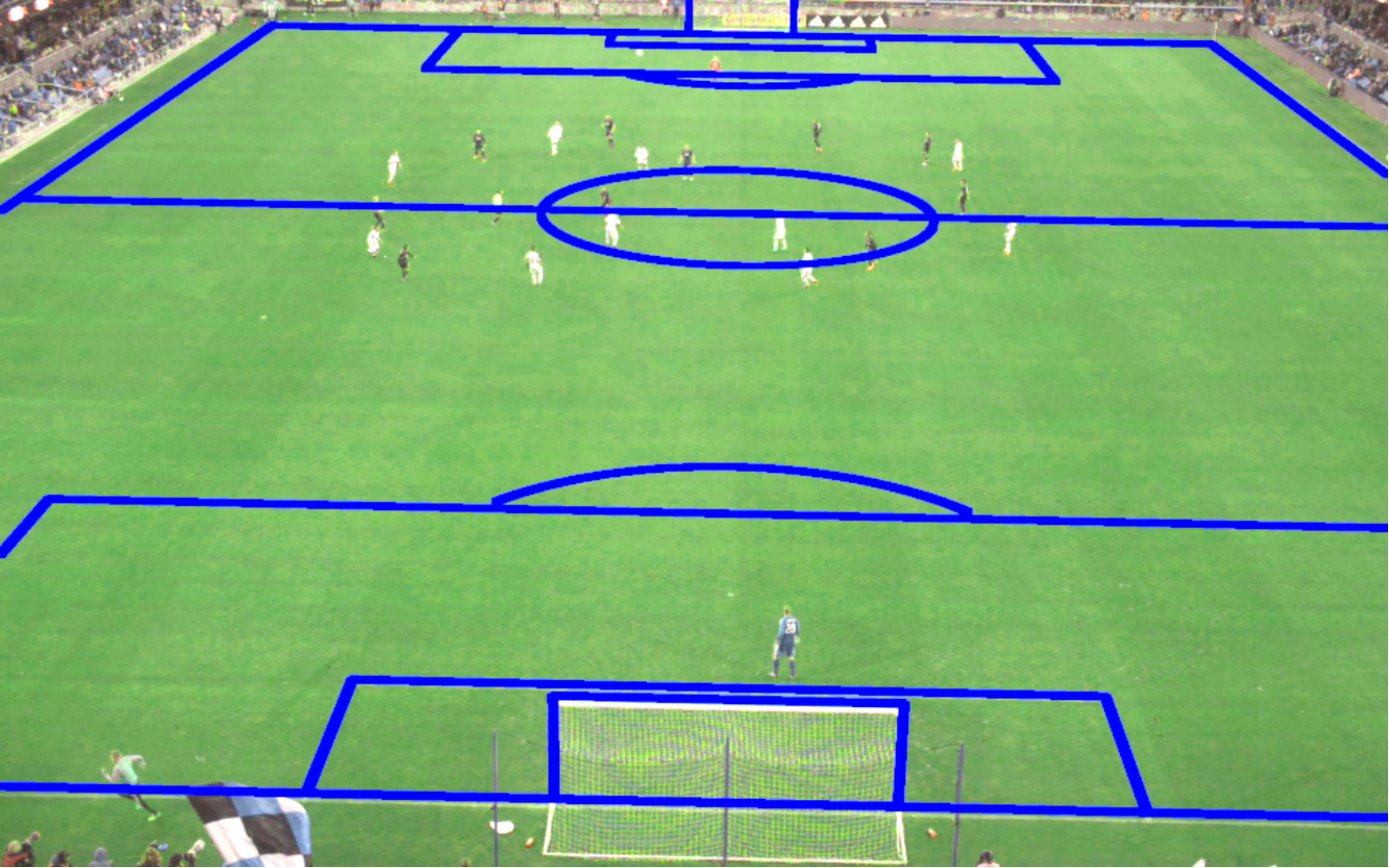} &
	    \includegraphics[width=.32\textwidth]{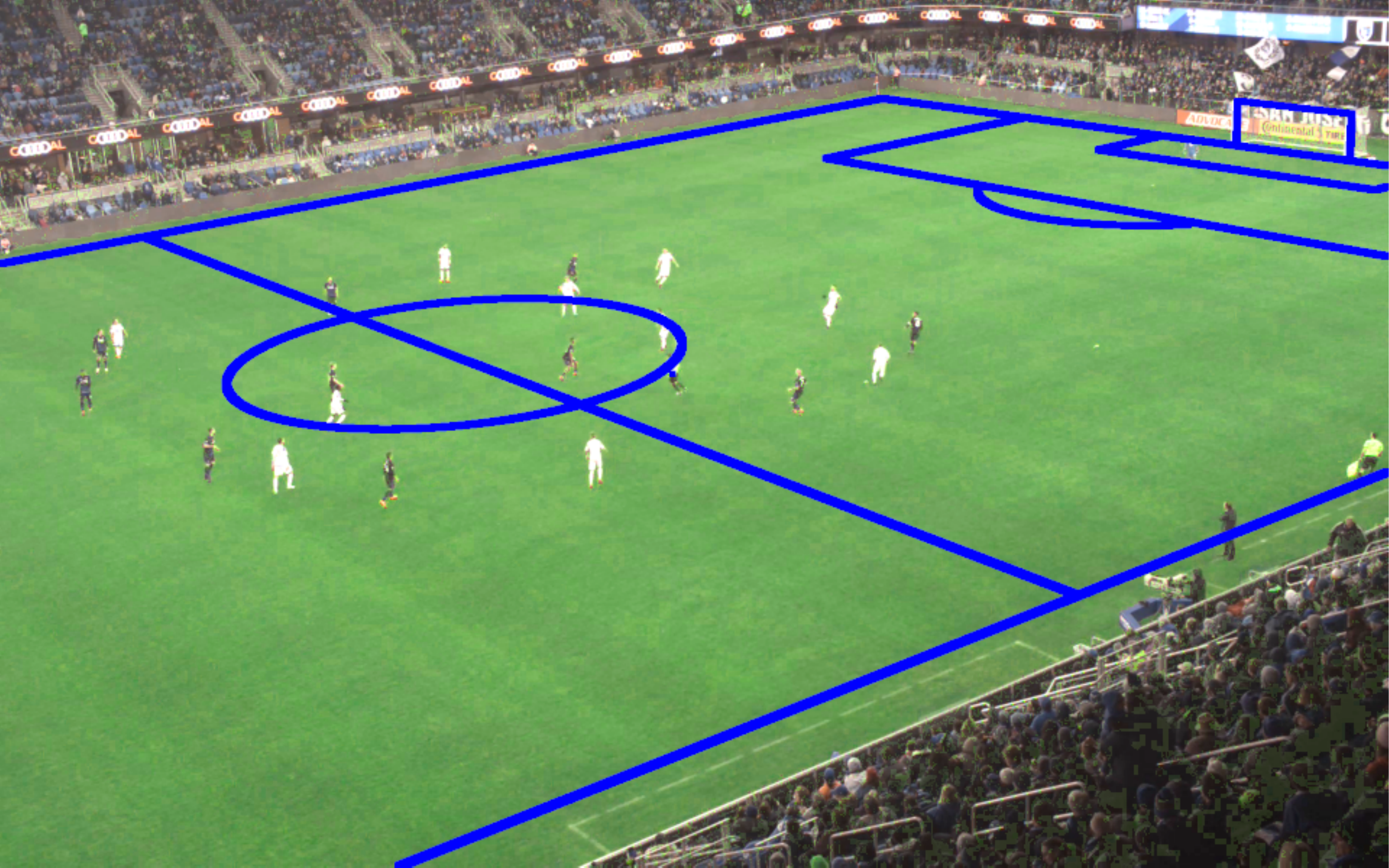} 
	  \end{tabular}
  }
  \caption{\small \textbf{Qualitative results.} 3D field lines projected and overlaid on the images according to the recovered camera registration.\label{fig:good_cases}}
  
\end{figure*}

%% file: fig_failure_cases.tex

\begin{figure*}[h!]
  \centering
  \setlength{\tabcolsep}{1pt} 
  \renewcommand{\arraystretch}{0.2} 
  \resizebox{\linewidth}{!}{%
	   \begin{tabular}{cccc}  
	     \includegraphics[width=.24\linewidth]{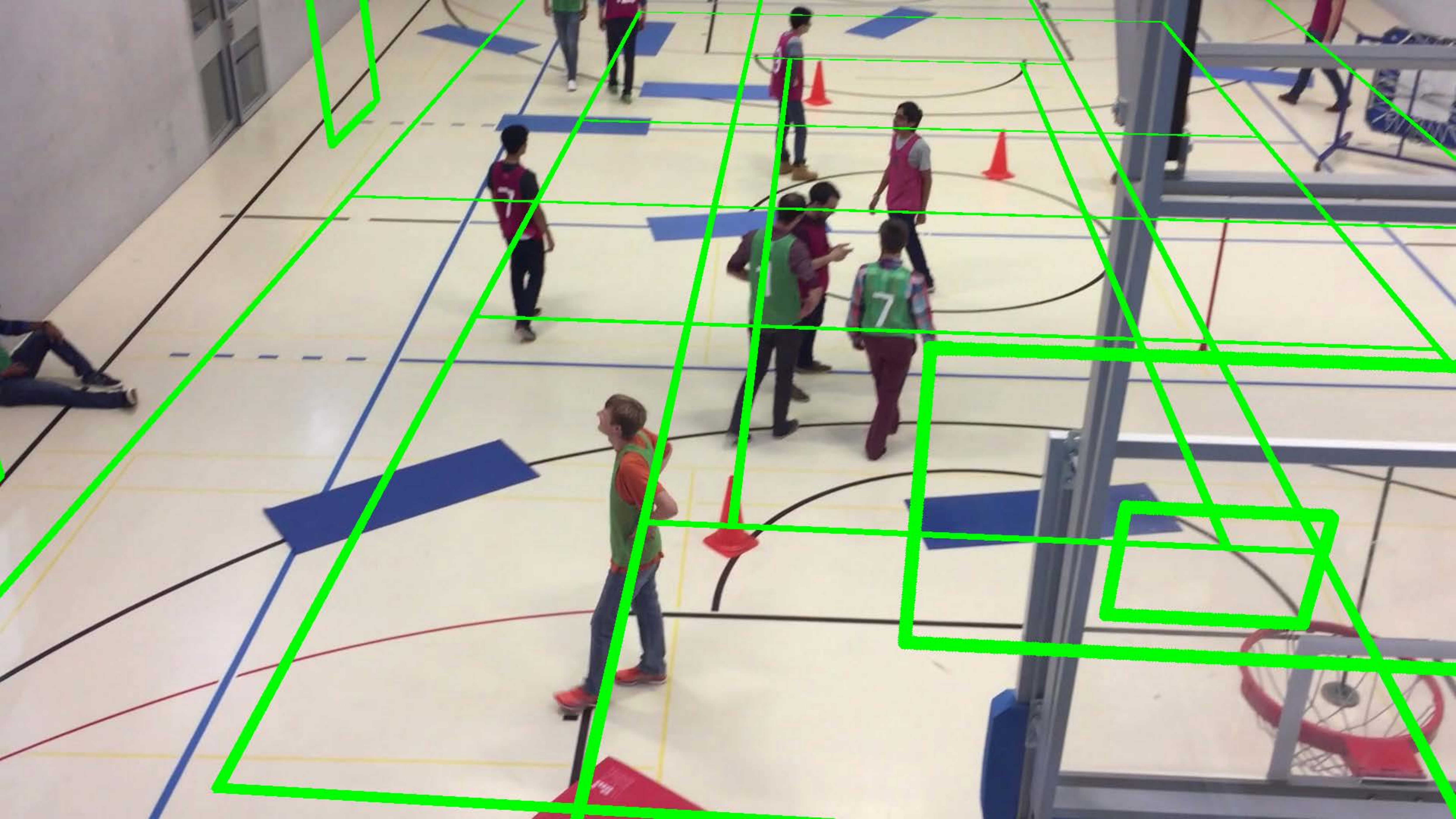} & 
	     \includegraphics[width=.24\linewidth]{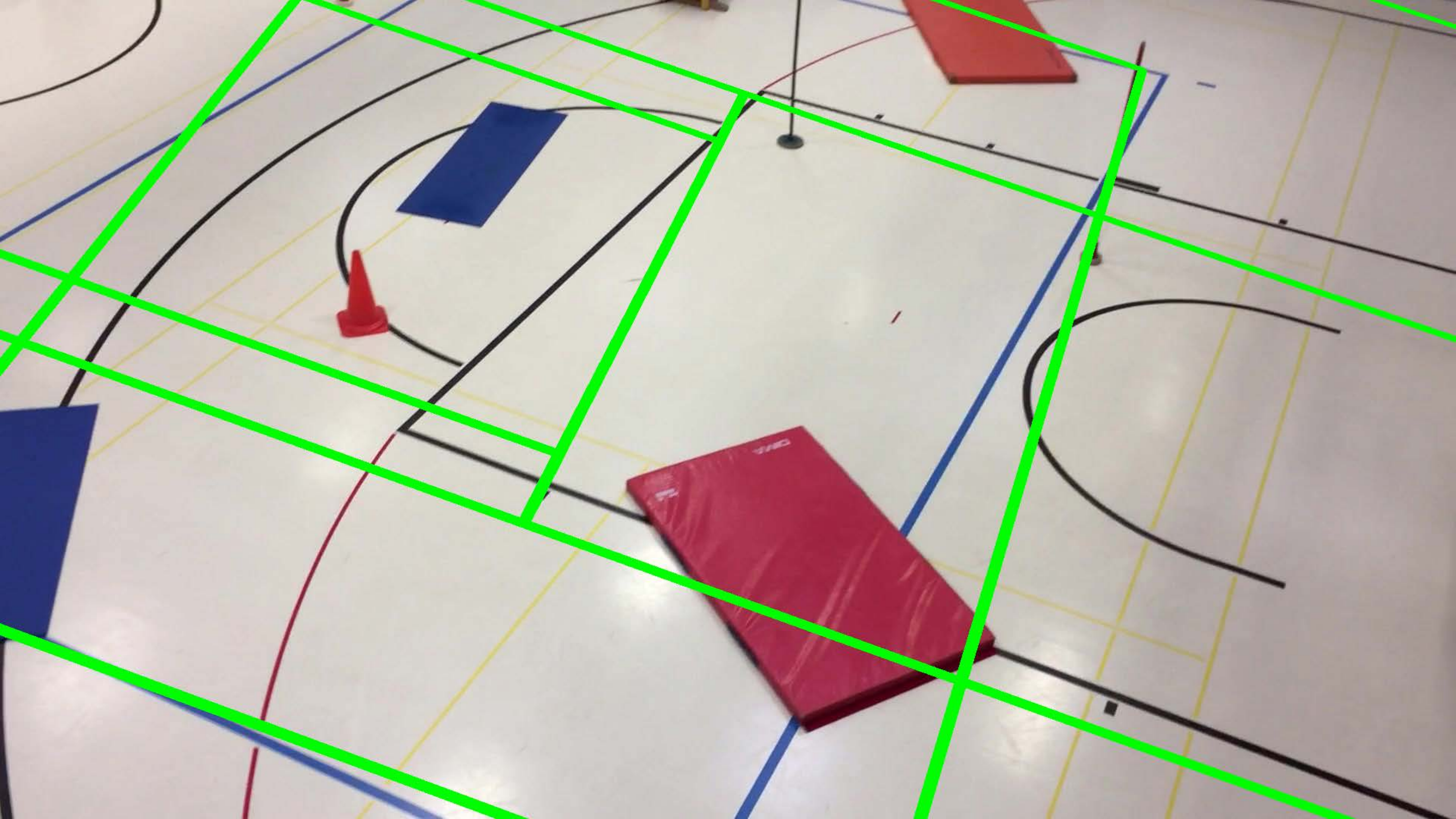} &
	     \includegraphics[width=.24\linewidth]{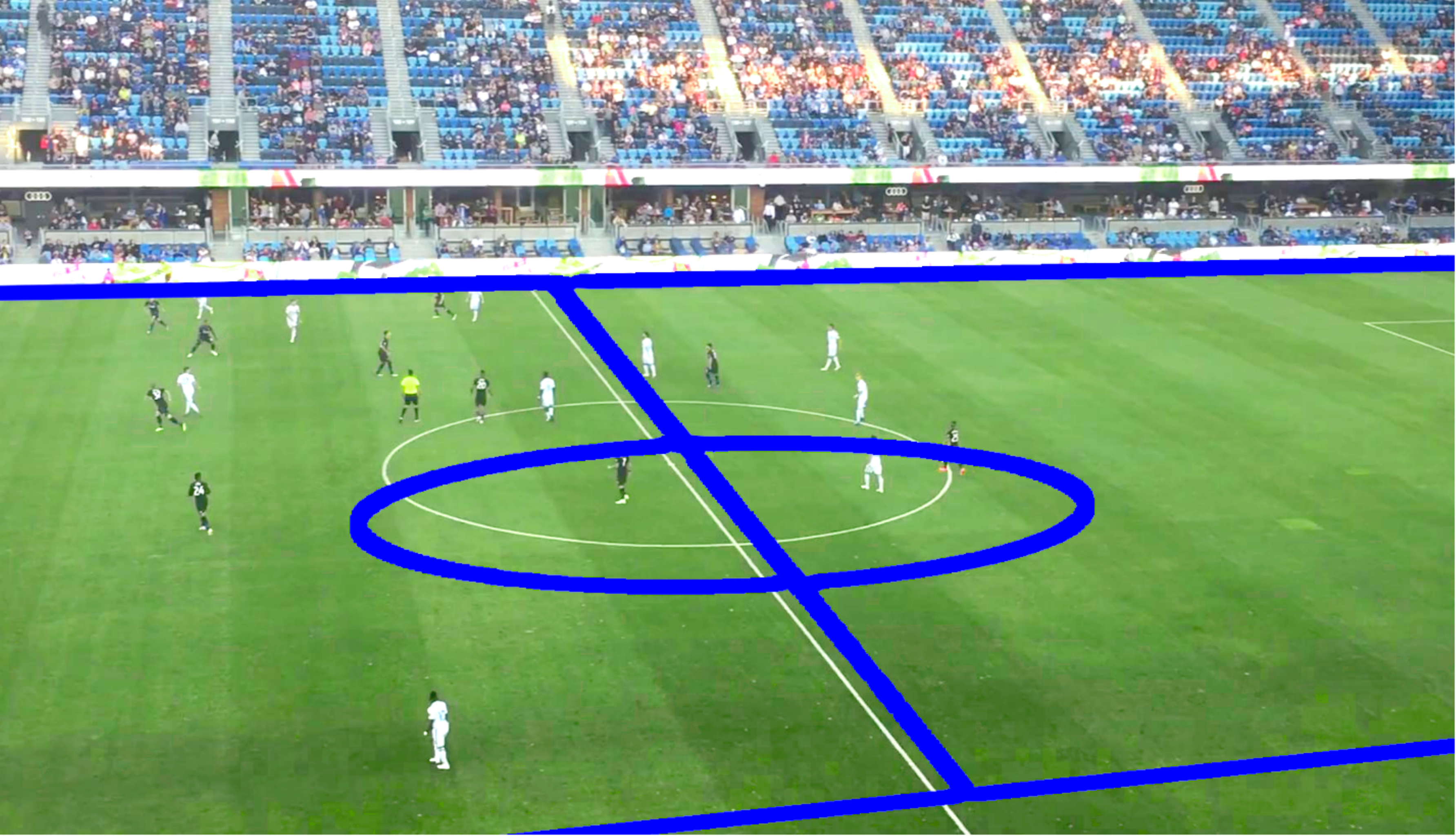} & 
	     \includegraphics[width=.24\linewidth]{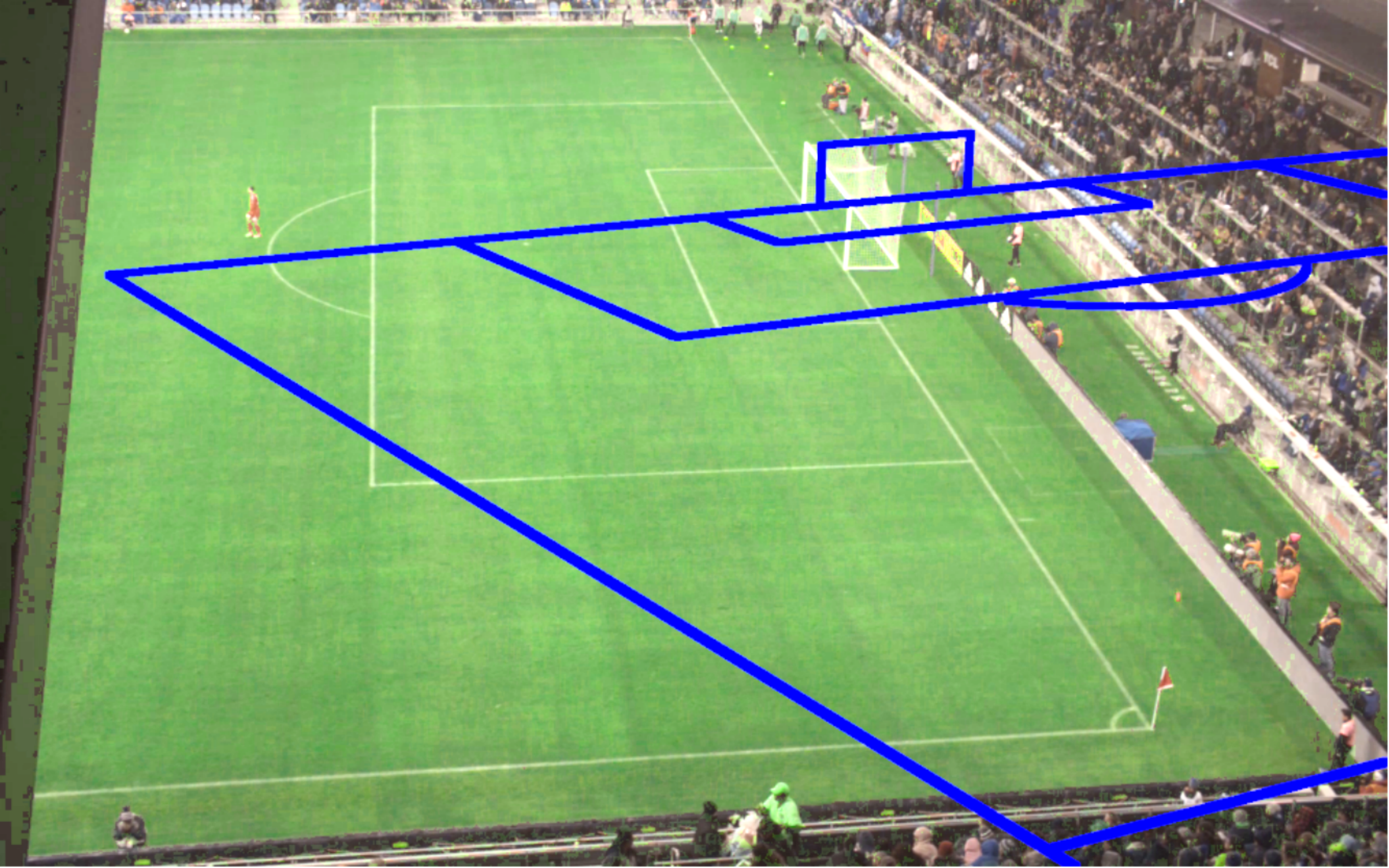}  \\  
	     (a) & (b) & (c) & (d)
	   \end{tabular}
 }     
	\caption{\small \textbf{Failure cases.} In the basketball dataset (a,b), narrow view points and clutter caused by the foreign objects, players and the hoop's frame are the main reasons for imprecise localization of keypoints leading to inaccurate poses. In (c), the inaccurate mapping is due to the shortage of visible lines. In (d), the network failed to locate keypoints correctly, this is most likely due to the fact that this viewpoint is far from the distribution of the training set we used. \label{fig:failure_cases}} 
\end{figure*}

%% file: fig_plots.tex

\begin{figure*}[h!]
\centering
  \begin{tabular}{@{}c | c | c | c@{}}
    \includegraphics[width=.23\textwidth]{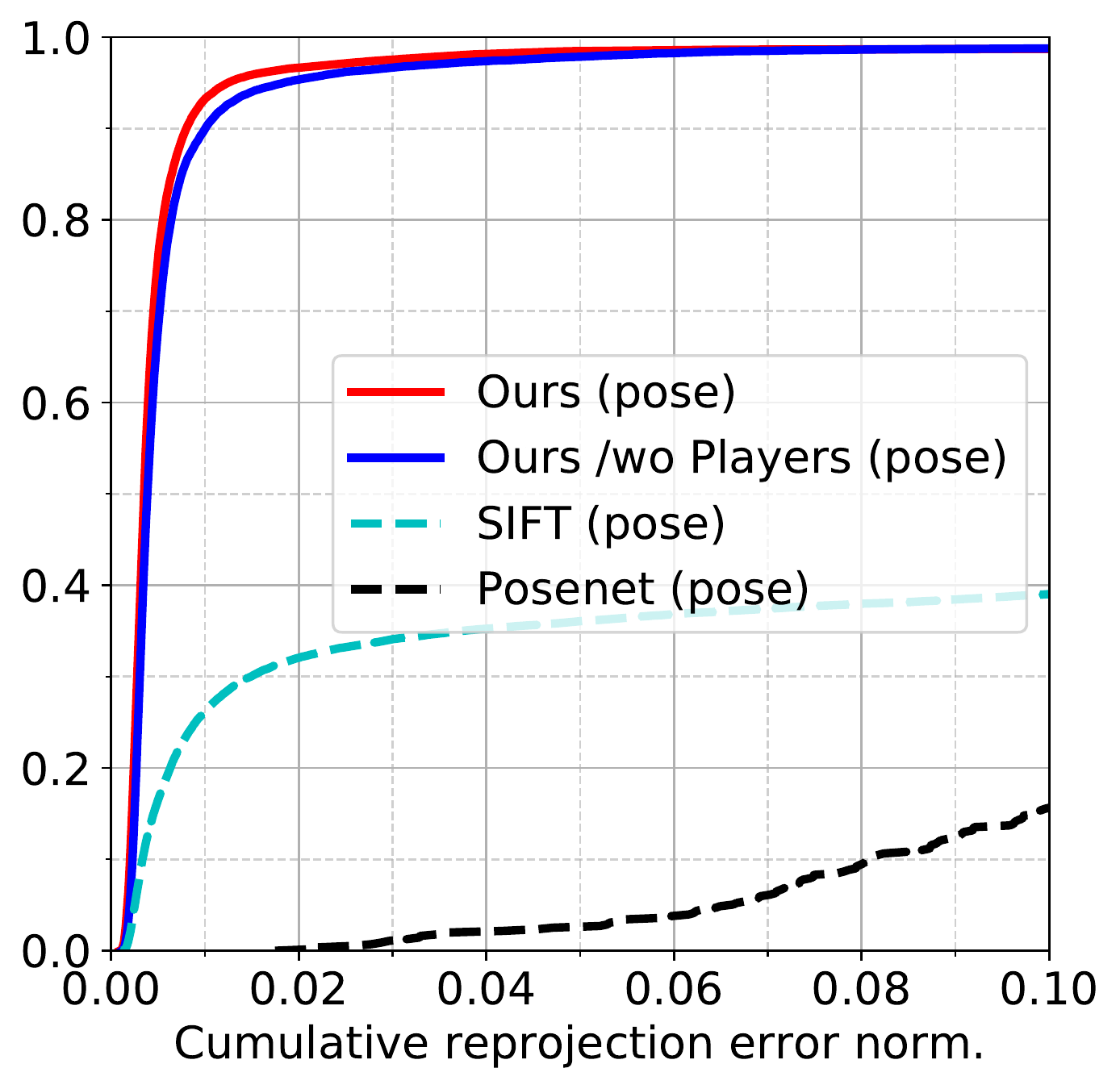}  &
    \includegraphics[width=.23\textwidth]{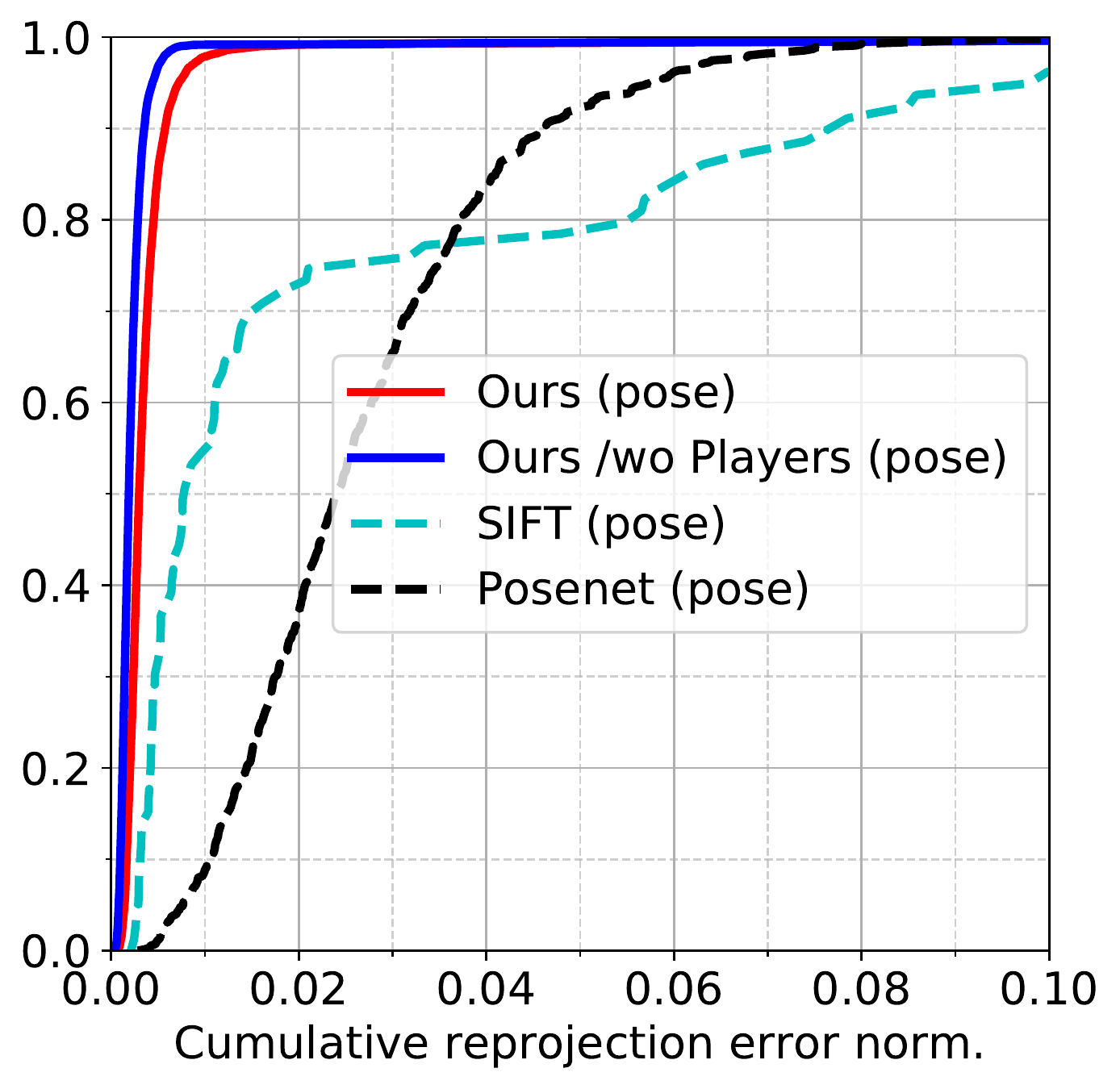}   &
    \includegraphics[width=.23\textwidth]{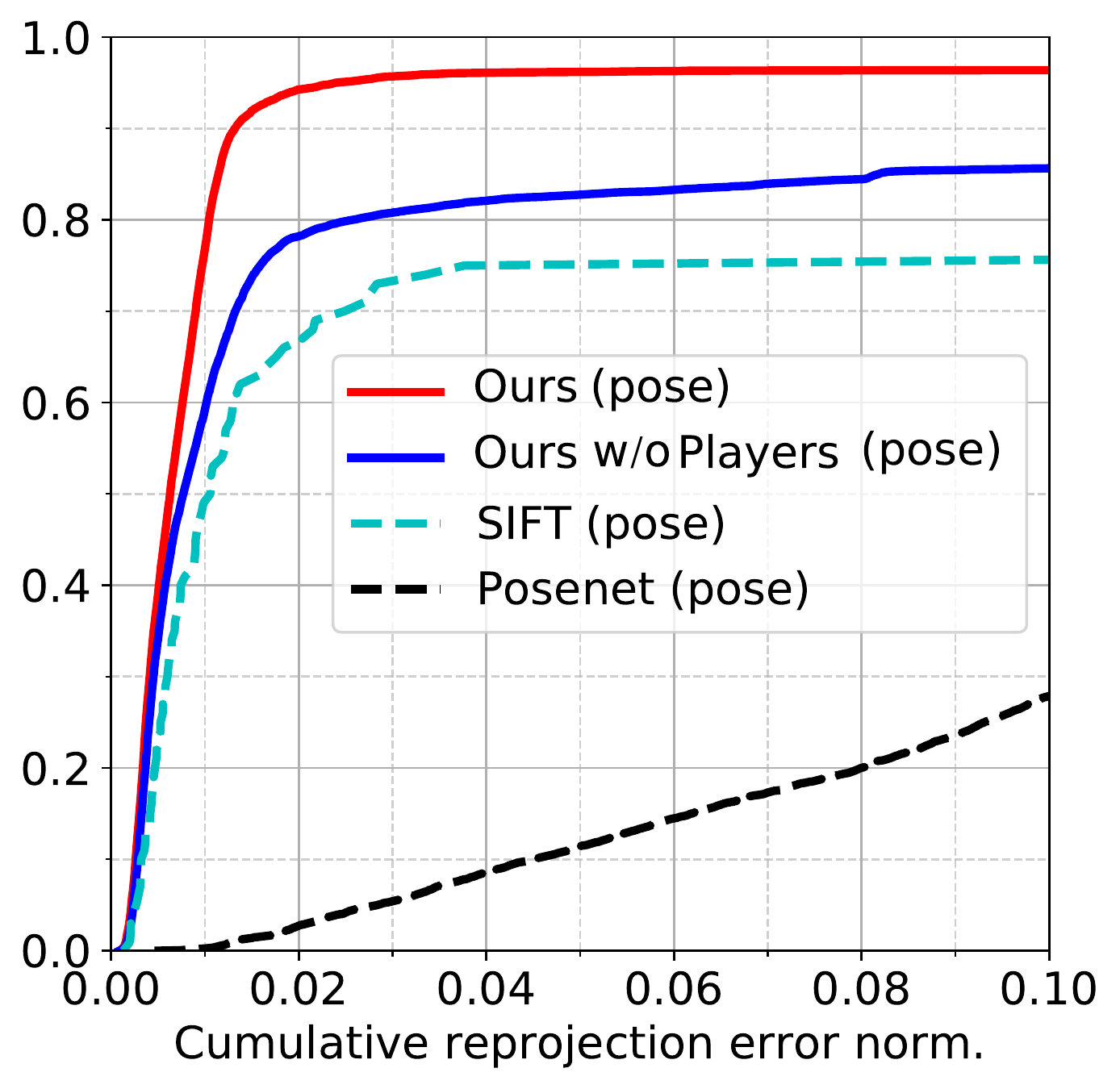} &
    \includegraphics[width=.23\textwidth]{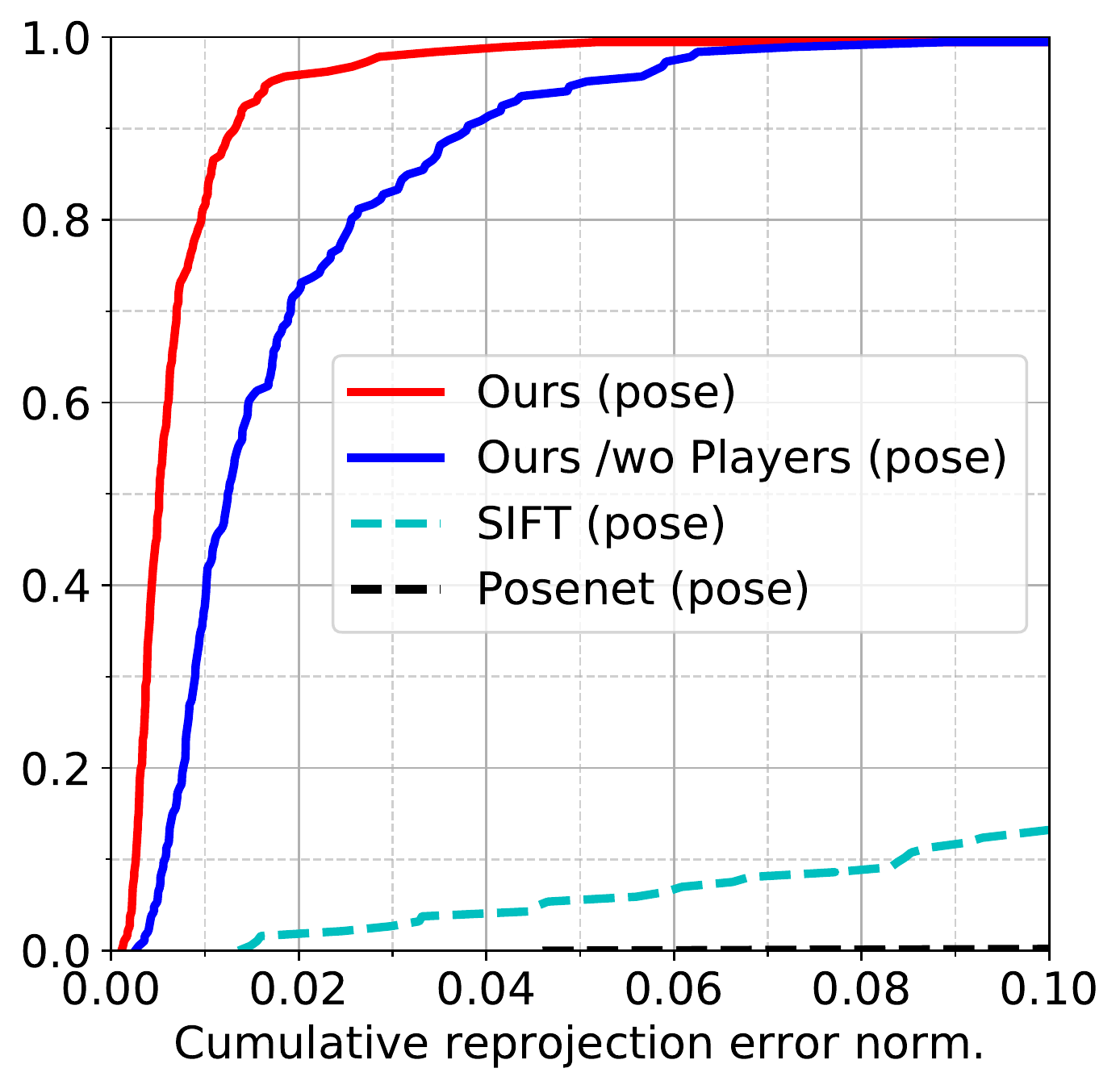}\\
    \includegraphics[width=.23\textwidth]{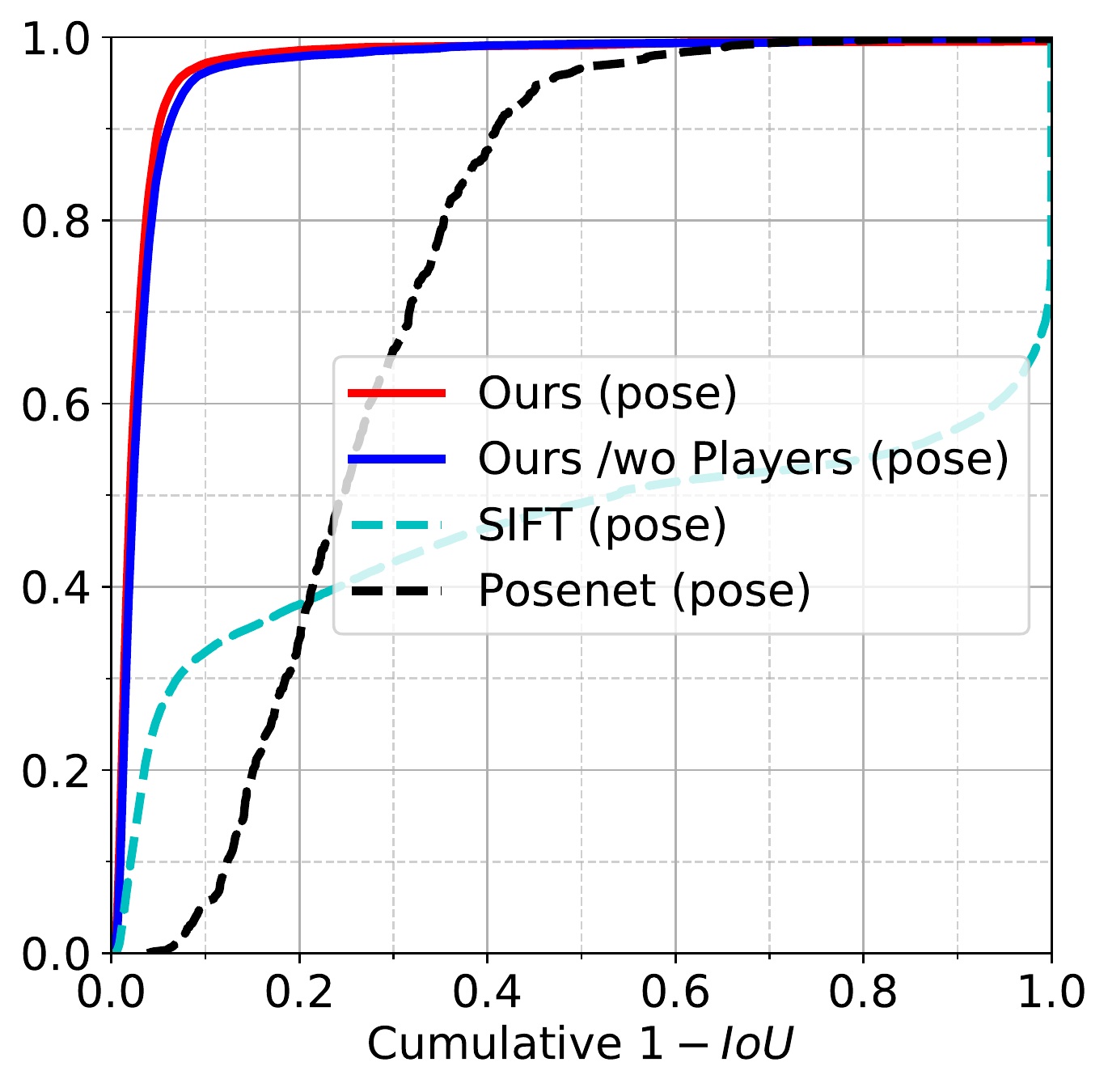} &
    \includegraphics[width=.23\textwidth]{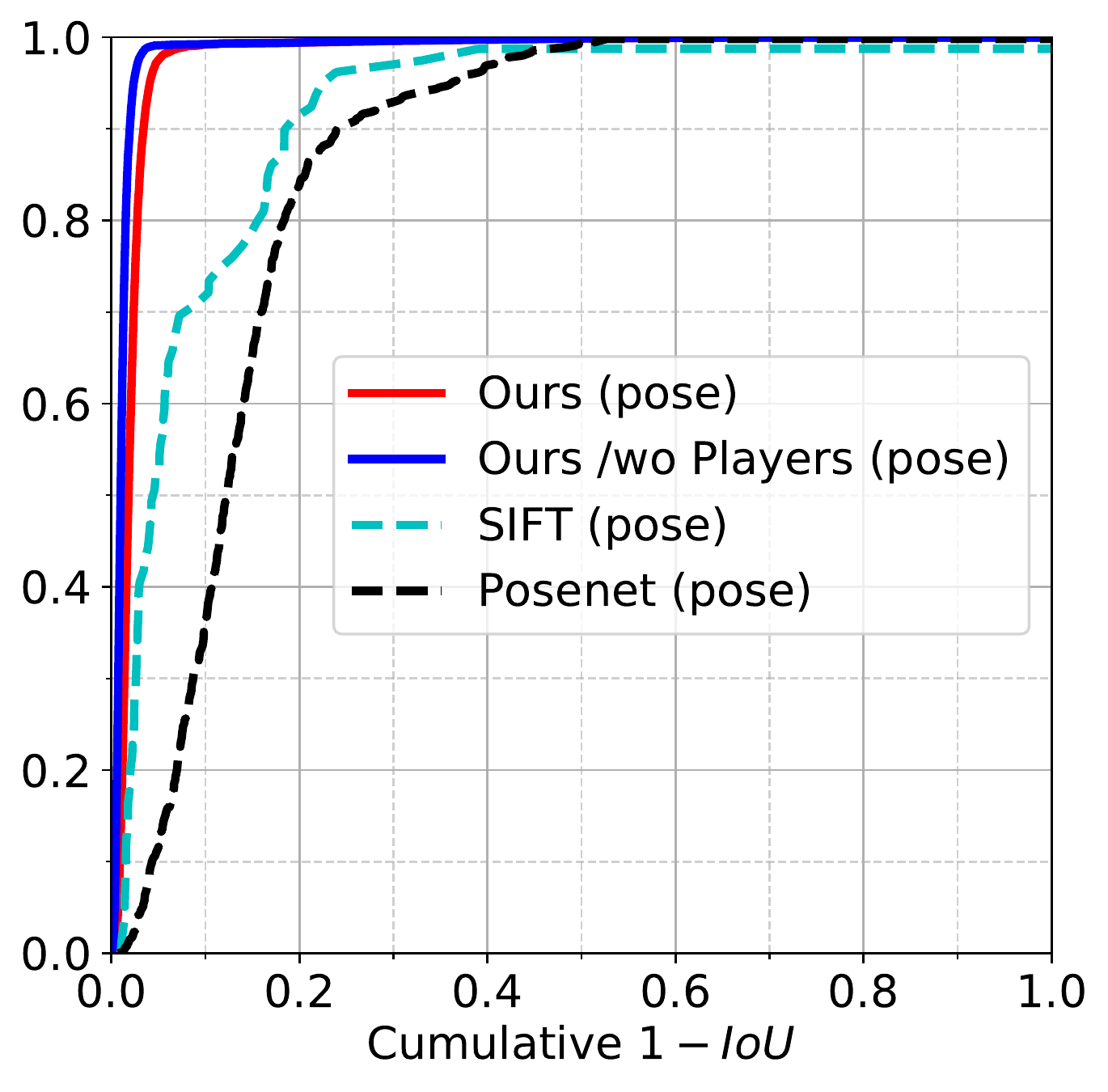} &
    \includegraphics[width=.23\textwidth]{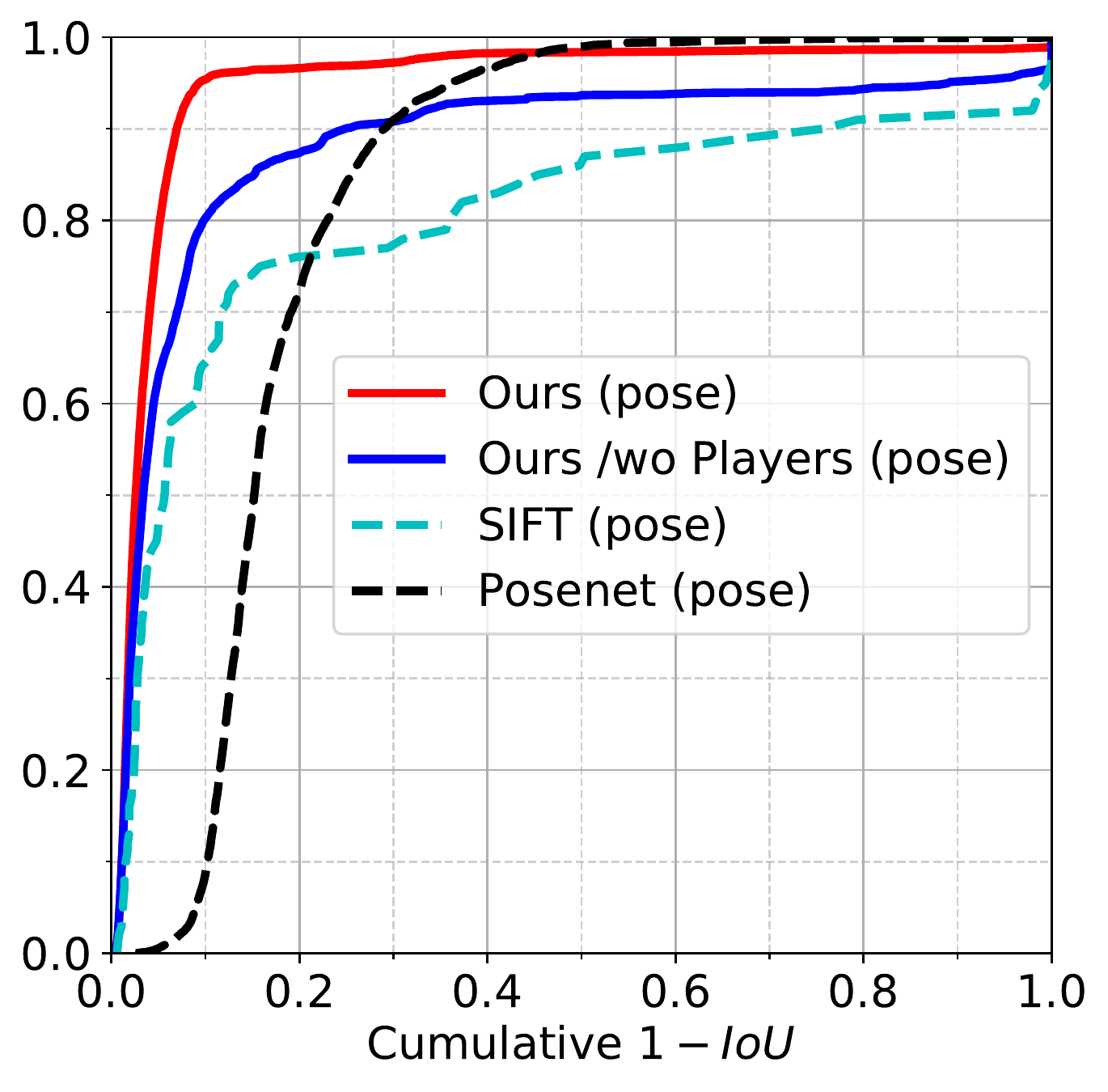} &
    \includegraphics[width=.23\textwidth]{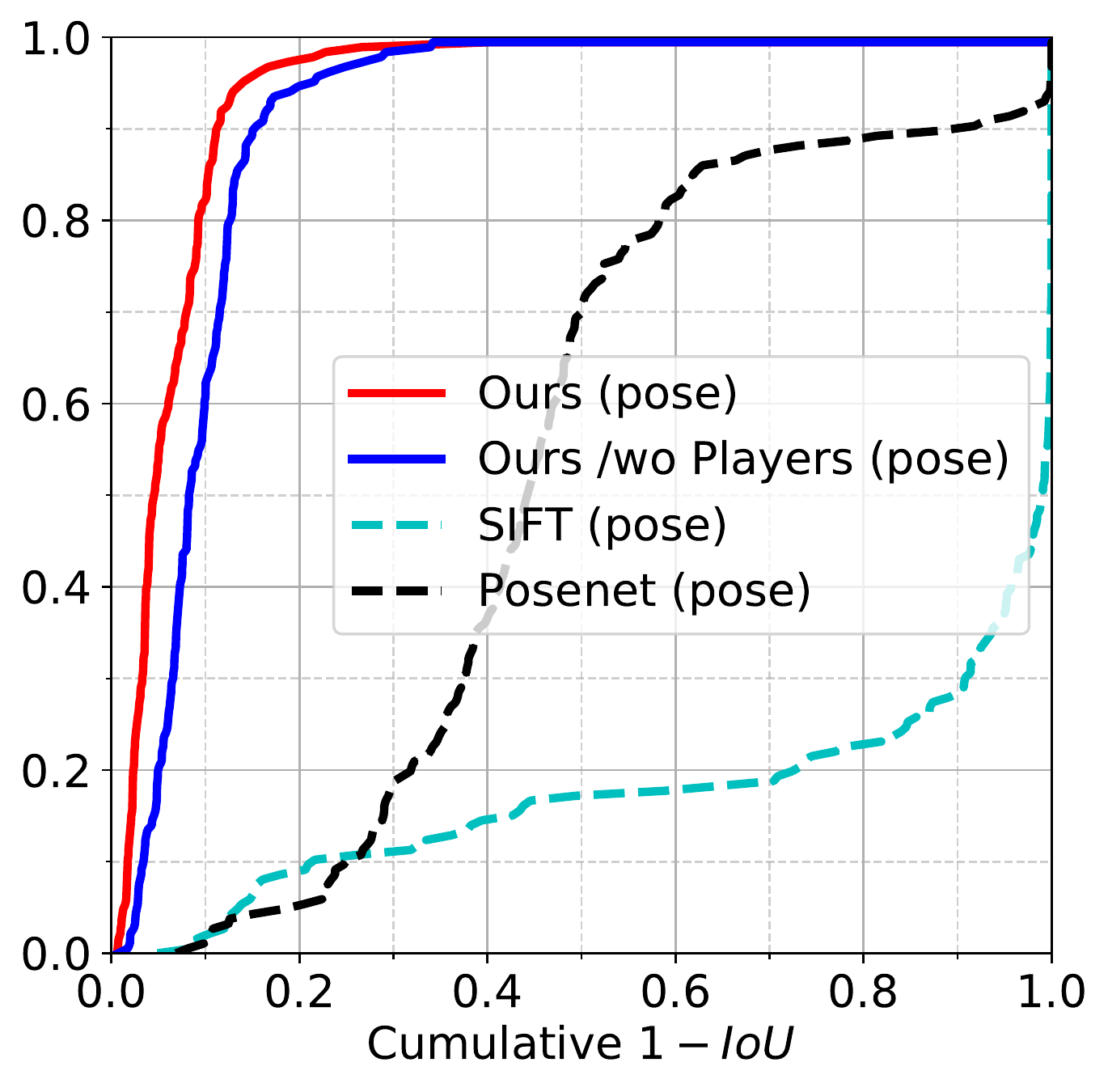}\\
    \basket & \volley & \soccerM & \soccerW
  \end{tabular}
  \caption{\textbf{Quantitative results.} Top row: cumulative distribution of normalized reprojection errors (NRE).
  Bottom row: cumulative distribution of one-minus intersection-over-union (1-IoU).
  For basketball and volleyball datasets the players data do not improve the accuracy of the estimation,
  by contrast, for both soccer datasets the players are key to top performance. \label{fig:plots}}
\end{figure*}

%% file: table_results_ours.tex

\begin{table*}[h!]
\caption{Quantitative results for \bbasket{}, \bvolley{} and \bsoccerM{} datasets. \label{tab:results_ours}}
\centering
\resizebox{\linewidth}{!}{%
	\setlength{\tabcolsep}{3pt}
	\begin{tabular}{llc|ccc|ccc|ccc|ccc|ccc}
	\toprule

& & \multicolumn{1}{c|}{\textbf{fps}}
& \multicolumn{3}{c|}{\textbf{IoU}}
& \multicolumn{3}{c|}{\textbf{Norm. reproj. error}}
& \multicolumn{3}{c|}{\textbf{Angular error (\(^{\circ
}\))}}
& \multicolumn{3}{c|}{\textbf{Translation error (m)}}
& \multicolumn{3}{c}{\textbf{Rel. focal length error}} \\

& & Mean
& Mean & Median & 1-AUC$^{<1}$
& Mean & Median & AUC$^{<0.1}$
& Mean & Median & AUC$^{<10}$
& Mean & Median & AUC$^{<2.5}$
& Mean & Median & AUC$^{<0.1}$\\

	\midrule
	\multirow{6}{*}{\rotatebox{90}{\bbasket}} &
	\ours                     &          22  &  \textbf{0.966}& \textbf{0.980}& \textbf{0.966}&         0.013 &  \textbf{0.003}& \textbf{0.940}&           1.681 &          0.565 &  \textbf{0.926}&         0.330 &         0.189 & \textbf{0.906}&         0.012 &         0.011 &         0.880 \\
	& \oursNOpl               &          26  &          0.962 &         0.977 &         0.962 & \textbf{0.012}&          0.004 &         0.930 &           1.179 &          0.610 &          0.917 &         0.294 &         0.194 &         0.898 &         0.013 &         0.012 &         0.867 \\
	& \oursNOpf               &  \textbf{33} &          0.927 &         0.978 &         0.927 &         0.041 &          0.004 &         0.895 &           6.962 &          0.590 &          0.884 &         1.865 &         0.196 &         0.865 &         0.012 &         0.011 &         0.879 \\
	& \oursNOsd               &          22  &          0.962 &         0.979 &         0.962 &         0.018 &          0.004 &         0.921 &   \textbf{1.107}&  \textbf{0.563}&          0.915 & \textbf{0.291}& \textbf{0.180}&         0.898 & \textbf{0.010}& \textbf{0.006}& \textbf{0.904} \\
	& \sift~\cite{Lowe04}     &         0.6  &          0.463 &         0.462 &         0.463 &         0.308 &          0.239 &         0.336 &          59.137 &         28.489 &          0.345 &         7.389 &         4.591 &         0.311 &         0.408 &         0.357 &         0.273 \\
	& \poseN~\cite{Kendall15a}&          19  &          0.739 &         0.755 &         0.738 &         0.385 &          0.284 &         0.046 &           5.709 &          3.743 &          0.559 &         3.319 &         2.645 &         0.226 &             - &           -   &            -  \\
	\midrule

	\multirow{6}{*}{\rotatebox{90}{\bvolley}} &
	\ours                     &  32          &         0.978 &         0.982 &         0.977 &         0.005 &         0.003 &         0.960  &         0.444 &         0.282 &         0.965 &         0.702 &         0.519 &         0.726 &         0.023 & \textbf{0.015}&         0.776 \\
	& \oursNOpl               &  38          & \textbf{0.987}& \textbf{0.990}& \textbf{0.987}& \textbf{0.003}& \textbf{0.002}& \textbf{0.973} & \textbf{0.284}& \textbf{0.177}& \textbf{0.976}&         0.662 & \textbf{0.491}&         0.742 & \textbf{0.021}& \textbf{0.015}& \textbf{0.788}\\
	& \oursNOpf               &  \textbf{43} &         0.976 &         0.979 &         0.976 &         0.004 &         0.004 &         0.957  &         0.424 &         0.339 &         0.957 &         0.711 &         0.525 &         0.721 &         0.023 &         0.016 &         0.773 \\
	& \oursNOsd               &  32          &         0.976 &         0.981 &         0.976 &         0.005 &         0.003 &         0.957  &         0.462 &         0.311 &         0.961 &         0.863 &         0.604 &         0.671 &         0.028 &         0.020 &         0.722 \\
	& \sift~\cite{Lowe04}     &   2          &         0.923 &         0.957 &         0.913 &         0.024 &         0.008 &         0.770  &         1.587 &         0.513 &         0.838 &         2.792 &         0.800 &         0.506 &         0.081 &         0.026 &         0.613 \\
	& \poseN~\cite{Kendall15a}&  19          &         0.861 &         0.877 &         0.859 &         0.027 &         0.024 &         0.731  &         0.615 &         0.568 &         0.937 & \textbf{0.556}&         0.499 & \textbf{0.777}&             - &             - &             - \\
	\midrule

	\multirow{6}{*}{\rotatebox{90}{\bsoccerM}} &
	\ours                     &          21  &  \textbf{0.949}&         0.974 & \textbf{0.948}& \textbf{0.021}& \textbf{0.006}& \textbf{0.898}& \textbf{3.973}&         0.927 & \textbf{0.864}&         4.895 &         2.668 & \textbf{0.192}& \textbf{0.035}&         0.008 &         0.833  \\
	& \oursNOpl               &          25  &          0.885 &         0.966 &         0.885 &         0.055 &         0.008 &         0.769 &         7.985 &         1.265 &         0.754 &         9.756 &         2.833 &         0.181 &         0.112 &         0.011 &         0.727  \\
	& \oursNOpf               &  \textbf{31} &          0.923 &         0.970 &         0.923 &         0.036 &         0.008 &         0.858 &         6.022 &         1.006 &         0.841 &         9.767 &         2.672 &         0.181 &         0.038 &         0.009 &         0.827  \\
	& \oursNOsd               &          21  &          0.943 & \textbf{0.976}&         0.942 &         0.030 & \textbf{0.006}&         0.881 &         7.070 & \textbf{0.898}&         0.848 & \textbf{4.813}& \textbf{2.509}&         0.190 &         0.046 & \textbf{0.007}& \textbf{0.836} \\
	& \sift~\cite{Lowe04}     &         0.8  &          0.809 &         0.944 &         0.804 &         0.137 &         0.010 &         0.680 &        12.553 &         1.104 &         0.732 &        15.223 &         3.059 &         0.146 &         0.128 &         0.013 &         0.709  \\
	& \poseN~\cite{Kendall15a}&          19  &          0.822 &         0.848 &         0.822 &         0.211 &         0.154 &         0.118 &        11.249 &         1.835 &         0.730 &        10.661 &         4.070 &         0.093 &             - &             - &             -  \\

	\bottomrule
	\end{tabular}
} 
\end{table*}

%% file: table_results_wc.tex

\begin{table}[h]
\caption{Quantitative results for \bsoccerW{} dataset. \label{tab:results_wc}
}
\centering
\resizebox{\linewidth}{!}{%
	\setlength{\tabcolsep}{3pt}
	\begin{tabular}{llc|ccc|ccc}
	\toprule

& & \multicolumn{1}{c|}{\textbf{fps}} 
& \multicolumn{3}{c|}{\textbf{IoU}} 
& \multicolumn{3}{c}{\textbf{Norm. reproj. error}}\\

& & Mean 
& Mean & Median & 1-AUC$^{<1}$ 
& Mean & Median & AUC$^{<0.1}$\\

	\midrule
	\multirow{7}{*}{\rotatebox{90}{\textbf{Soccer W. Cup}}} &	
	\ours                        &    8  &  \textbf{0.939}& \textbf{0.955}& \textbf{0.934}& \textbf{0.007}& \textbf{0.005}&  \textbf{0.926} \\
	& \oursNOpl                  &    9  &          0.905 &         0.918 &         0.901 &         0.018 &           0.012 &          0.820 \\
	& \sift~\cite{Lowe04}        &  1.6  &          0.170 &         0.011 &         0.168 &         0.591 &         0.479 &  0.01 \\
	& \poseN~\cite{Kendall15a}   &   19  &          0.528 &         0.559 &         0.525 &         0.849 &         0.878 &  0.00            \\	
	& \bbmrf~\cite{Homayounfar17}&  2.3  &          0.83  &            -  & - & - & - &     - \\
	& \hogdict~\cite{Sharma18}   &    5  &          0.914*&         0.927*& - & - & - &     - \\		
	& \errorref~\cite{Jiang19}   &   -   &          0.898 &         0.929 & - & - & - &     - \\
	\bottomrule
	\end{tabular}
	
}

\end{table}

%% file: table_keypoints_configs_table.tex

\begin{table}[h!]
\caption{\small \textbf{Testing different keypoint configurations.} We consider a detection to be an inlier if its distance to the closest corresponding ground-truth is less than 5 pixels in an image of $256 \times 455$ pixels. The mean distance is computed between the projected ground-truth keypoints and the inliers. \label{tab:keypointsConfigs}}
\centering
\resizebox{\linewidth}{!}{%
	\setlength{\tabcolsep}{3pt}
	\begin{tabular}{ l  c  c }
	\toprule
	\multirow{2}{*}{\textbf{Configuration}} & \textbf{Mean Distance} & \textbf{Inlier Proportion}\\  
	 & Mean$\pm$std. & Mean$\pm$std. \\ 
	 \cmidrule(r){1-1} \cmidrule(r){2-2} \cmidrule(r){3-3}
	Keypoints on corners 1 (red) & 1.15 $\pm$ 0.88 & 0.78 $\pm$ 0.22 \\ 
	Keypoints on corners 2 (blue) & 1.08 $\pm$ 0.85 & 0.79 $\pm$ 0.21 \\  
	Keypoints on a grid (white) & 1.66 $\pm$ 1.20 & 0.69 $\pm$ 0.27 \\  
	\bottomrule
	\end{tabular}
}
\end{table}

%% file: fig_keypoints_configs.tex

\begin{figure}[h!]
	\centering
	\includegraphics[width=\linewidth]{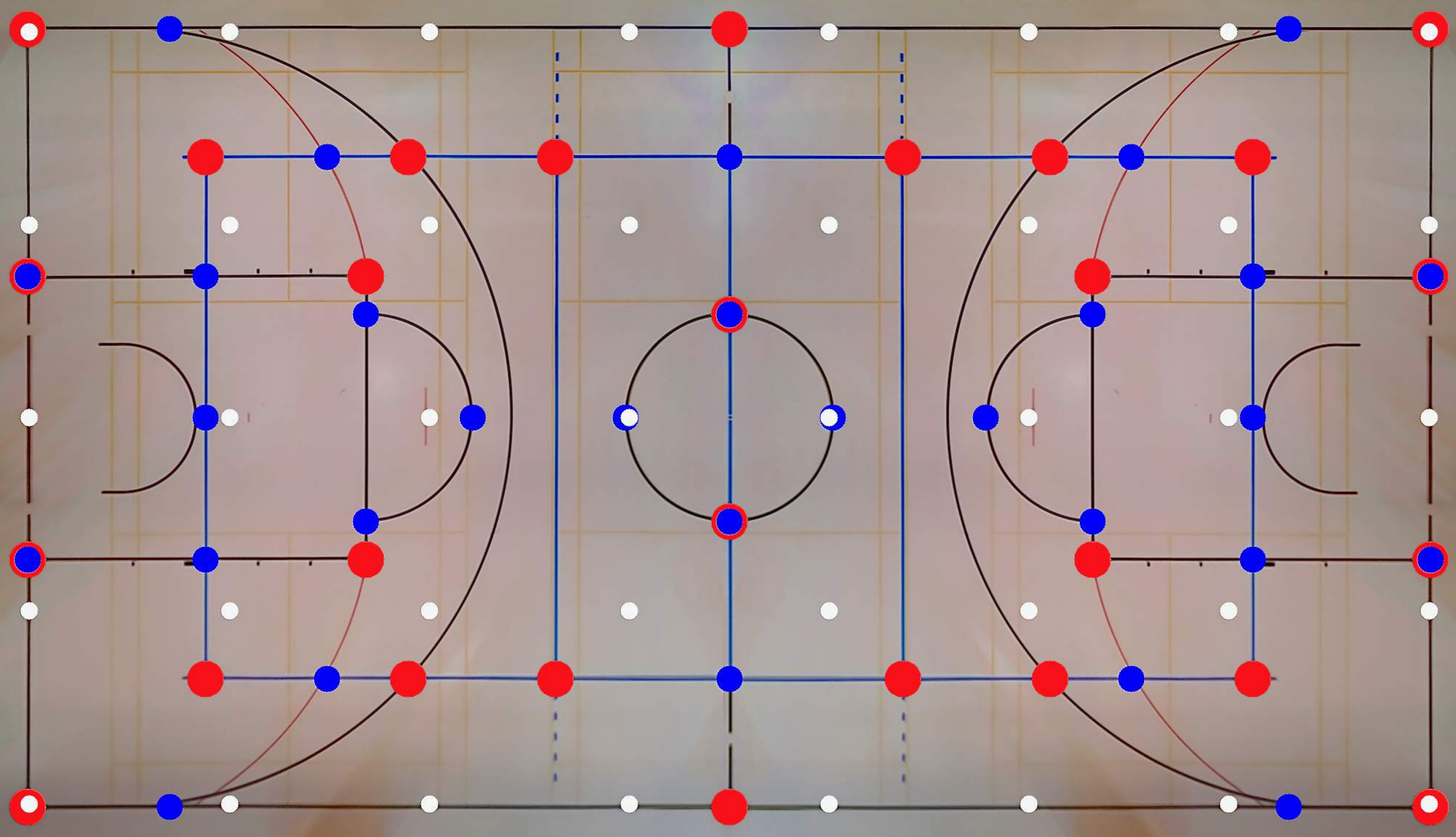}
	\caption{\small \textbf{Different keypoint configurations (\basket).} Red and blue dots depict two different configurations for semantic keypoints located at line intersections. The white ones are equally spaced and represent a third.\label{fig:keypointsConfigs}}
\end{figure}

%% file: 5_conclusion.tex

\section{Conclusion}

\blfootnote{*computed using the visible part of the field in the image. See Appendix~\ref{sec:iou_issue} for more explanations.}
We have developed a  new camera-registration framework that combines accurate localization and robust identification of specific keypoints in the image by using a fully-convolutional deep architecture. It derives its robustness and accuracy from being able to jointly exploit the information by the field lines and the players' 2D locations, while also enforcing temporal consistency. 

Future work will focus on detecting not only the 2D location of the projection of the players' center of gravity, as we currently do, but also their joints so that we can reconstruct their 3D pose. In this way, we will be able to simultaneously achieve camera registration and 3D pose estimation. This has a tremendous potential in terms of augmenting the images and developing real-time tools that could be used to explain to viewers what the action is. 

%% file: 7_acknowledgments.tex

\section{Acknowledgments}


This work was supported in part under an Innosuisse Grant funding the collaboration between SecondSpectrum and EPFL.

%% file: 6_appendix.tex

\vspace{1cm}
\begin{center}
{\Large \bf Appendix}
\end{center}

\section{From Homography to Camera Parameters}
\label{sec:cameraParams}

Let us consider a $w \times h$ image $I$ along with the homography~$\bH$ between ground plane and its image plane, which is to be decomposed into $\bK$ and~$\bM = \lbrack \bR,\bt \rbrack$, a $3 \times 3$ matrix of intrinsic parameters and a $3 \times 4$ matrix of extrinsic parameters, as defined in Section~\ref{sec:formal}.  In this section, we outline how to derive $\bM$ and $\bK$ from $\bH$. For a full treatment, we refer the interested reader to~\cite{Hartley00}. 

\parag{Intrinsic parameters.}
In practice, the principal point of modern cameras is located close to the center of the image and there is no skew. We can therefore write 
\begin{align}
\bK &= \begin{bmatrix}
       f & 0 & w/2  \\[0.3em]
       0 & f & h/2  \\[0.3em]
       0         & 0 & 1
       \end{bmatrix} \; ,
\label{eq:intrinsic}
\end{align}
where $f$ is the initially unknown focal length and the only parameter to be estimated. It can be shown that knowing $\bH$, two linear constraints on the intrinsics parameters can be solved for the unknown $f$ which yield two solutions of the form:
\begin{align}
f_1 & = \frac{g_1(\mathbf{h_1}, \mathbf{h_2},w,h)}{h_7 \cdot h_8}                 \label{eq:focalLs} \; ,                  \\
f_2 & = \frac{g_2(\mathbf{h_1}, \mathbf{h_2},w,h)}{(h_7 + h_8)\cdot(h_7 - h_8)} \nonumber \; ,
\end{align}
where $\mathbf{h_1}$ and $\mathbf{h_2}$ are the first two columns of $\bH$, $h_7$ and $h_8$ the first two elements of $\bH$ third row, and $g_1$, $g_2$ are algebraic functions. 

$f_1$ and $f_2$ are only defined when the denominators are non-zero and the closer to zero they are, the less the precision. In practice we compare the value of these denominators and use the following heuristic
 \begin{equation}
f =
  \begin{cases}
        f_1   & \quad |h_7 \cdot h_8| > |(h_7 + h_8)\cdot(h_7 - h_8)| \; . \\
       f_2  & \quad \mbox{otherwise.}
  \end{cases}
\label{eq:focalL}
\end{equation}

\parag{Extrinsic parameters.}
To extract the  rotation and translation matrices $\bR$ and $\mathbf{t}$ from $\bH$, we first define the $3 \times 3$ matrix $\mathbf{B}=\lbrack \mathbf{b_1}, \mathbf{b_2}, \mathbf{b_3} \rbrack$ and a scale factor $\lambda$ to write  $\bH$ as $\lambda \bK \mathbf{B}$.  $\lambda$ can be computed as $(||\bK^{-1} \mathbf{h_1} ||+||\bK^{-1} \mathbf{h_2} ||)/2$. Then, assuming that the x-axis and y-axis define the ground plane, we obtain a first estimate of the rotation and translation matrices $\tilde{\bR} = \lbrack \mathbf{b}_1, \mathbf{b}_2, \mathbf{b_1} \times \mathbf{b_2} \rbrack$ and $\mathbf{t}=\mathbf{b_3}$. We orthogonalize the rotation using singular value decomposition $\mathbf{\tilde{R}} = \mathbf{U} \bf \Sigma \mathbf{V^T}, \ \bR = \mathbf{U}\mathbf{V^T}$. Finally, we refine the pose $\lbrack \bR,\mathbf{t} \rbrack$ on $\bH$ by non-linear least-squares minimization.

\section{Complete Framework}
\label{sec:framework}

Recall from Section~\ref{sec:keypoints} that at each discrete time step $t$, we estimate the 2D locations of our keypoints $\hat \bz^t$, which are noisy and sometimes plain wrong. 
As we have seen in Section~\ref{sec:cameraParams}, they can be used to estimate the intrinsic and extrinsic parameters $\bM_d^t$ and $\bK^t$ for single frames. The intrinsic parameters computed from a single frame are sensitive to noise and depend on the accuracy of $\bH^t$, for this reason, at every time step $t$ we estimate their values by considering past $k$ frames. We perform outlier rejection over the past $k$ estimate of the intrinsics then, compute the median, this allows to increase robustness and precision admitting smooth variations of the parameters over time. If the parameter are known to be constant over time, $k$ can be set so to consider all past estimates. Once the intrinsics are computed, we obtain the new robust pose $\bM^t$ from the filter and minimize the error in the least-squares sense using all the detected keypoints.

This particle-filter is robust but can still fail if the camera moves very suddenly. To detect such events and re-initialize it, we keep track of the number of 3D model points whose reprojection falls within a distance $t$ for the pose computed from point correspondences $\hat \bM_d^t$ and the filtered pose $\hat \bM^t$. When the count for $\hat \bM_d^t$ is higher we re-initialize the filter. 

The pseudo code shown in {\bf Algorithm}~\ref{fig:pseudocode} summarizes these steps.

\input{pseudocode_framework}

\section{Intersection-Over-Union (IoU) of the visible area}
\label{sec:iou_issue}
In~\cite{Sharma18}, the intersection-over-union metric is computed using only the area of the court that is visible in the image. This area is shown in gray in Figure~\ref{fig:iou_issue}. After superimposition of the projected model (red frame) with the ground-truth one (blue frame) the area of the court that is not visible in the image is removed and therefore not taken into account in the computation of the IoU. It can be shown that the IoU of the gray area gives a perfect score while in reality the estimate is far from correct. The worst case scenario is when the viewpoint leads to an image containing only grassy area of the playing field. In this case, as long as the projected model covers the ground-truth one, this metric gives perfect score. For this reason, we discourage the use of this version of IoU.

\input{fig_iou_issue.tex}

%% file: pseudocode_framework.tex

\begin{algorithm}
\scriptsize{
\begin{algorithmic}[1]
\Procedure{Intrinsics and extrinsics estimation}{}
\item \textbf{for} $t=1$ \textbf{to} T: iterates over time
\item \quad -----Single frame estimation-----
\item \quad $\hat \bz^t=\{\hat \bz^t_S, \hat \bz^t_P\} \leftarrow$ detect keypoints from $I^t$
\item \quad $\hat \bH^t \leftarrow$ robust estimation using ($\bZ_S, \hat \bz^t_S$)
\item \quad $\hat \bH^t_r \leftarrow$ refinement using Players ($\hat \bH^t, \bZ^t,\hat \bz^t$)
\item \quad $\hat \bK^t \leftarrow$ intrinsics estimation from $\hat \bH_r^t$
\item \quad $\hat \bK^t_m \leftarrow$ moving median /w outliers rejection over $\hat \bK^{t:t-k}$
\item \quad $\hat \bM_d^t \leftarrow$ homography decomposition $(\hat \bK^t_m, \hat \bH_r^t)$
\item \quad -----Particle Filtering-----
\item \quad \textbf{for} $n=1$ \textbf{to} N: iterates over the particles
\item \quad \quad $\{\bs^t_n,\pi^t_n\} \leftarrow$ sampling with replacement from $\{\bs^{t-1}_n,\pi^{t-1}_n\}$
\item \quad \quad $\bs^t_n \leftarrow \bs^t_n + w_n$ add randomness where $w_n \sim \mathcal{N}(0,\bf \Sigma)$
\item \quad \quad $\pi^{n} \leftarrow g(\bs^t_n, \hat \bK^t_m,\bZ_S, \hat \bz_S^t, \bZ_P^t, \hat \bz_P^t)$ weights computation
\item \quad \textbf{endfor}
\item \quad $\hat \bM^t \leftarrow \sum_{n=1}^{N} \pi_n^{t}\bs_n^t$ expected value as filter output
\item \quad $\hat \bM^t \leftarrow$ Levenberg–Marquardt refinement of $\hat \bM^t$ on $\hat \bH^t_r$
\item \quad -----Filter re-initialization-----
\item \quad \textbf{if} no. inliers$(\hat \bK^t_m, \hat \bM_d^t, \hat \bz^t)>$ no. inliers$(\hat \bK^t_m, \hat \bM^t, \hat \bz^t)$ \textbf{then:}
\item \quad \quad \textbf{for} $n=1$ \textbf{to} N:
\item \quad \quad \quad $\{\bs^t_n, \pi^t_n\} \leftarrow \{\hat \bM_d^t + w_n, 1/N\}$ where $w_n \sim \mathcal{N}(0,\bf \Sigma)$
\item \quad \quad \textbf{endfor}
\item \quad \textbf{endif}
\EndProcedure
\end{algorithmic}
}
\caption{\small Complete framework pseudo code. \label{fig:pseudocode}}
\end{algorithm}

%% file: fig_iou_issue.tex

\begin{figure}[t]
\centering
    \resizebox{\columnwidth}{!}{%
	 \includegraphics{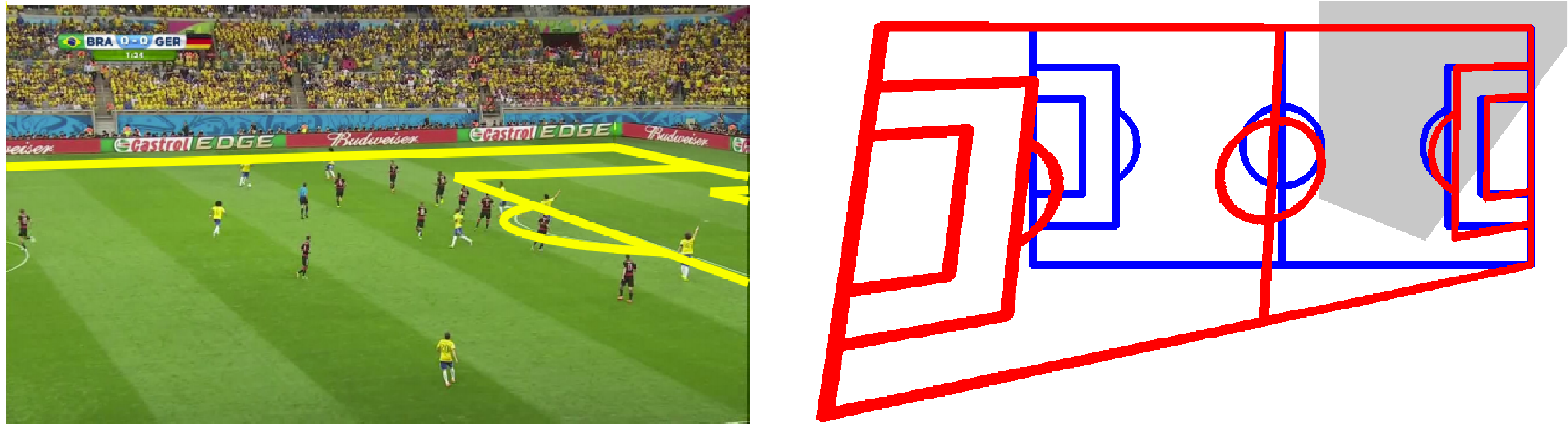}   
	 }
	\caption{Example failure case of the Intersection-over-union metric that only uses the visible part of the court in the image. The ground-truth model is shown in blue, the re-projected one in red and the gray area is the projected image plane. This version of IoU would give perfect score while the one that takes the whole template into account would give around $0.6$.\label{fig:iou_issue}}
\end{figure}